\documentclass[conference]{IEEEtran}
\IEEEoverridecommandlockouts
\usepackage{cite}
\usepackage{amsmath,amssymb,amsfonts}
\usepackage{algorithmic}
\usepackage{graphicx}
\usepackage{textcomp}
\usepackage{xcolor}
\usepackage{mathtools}
\usepackage{booktabs}
\usepackage{url}            % simple URL typesetting
\usepackage{subfig}

\def\BibTeX{{\rm B\kern-.05em{\sc i\kern-.025em b}\kern-.08em
    T\kern-.1667em\lower.7ex\hbox{E}\kern-.125emX}}

\newcommand\blfootnote[1]{%
  \begingroup
  \renewcommand\thefootnote{}\footnote{#1}%
  \addtocounter{footnote}{-1}%
  \endgroup
}

\DeclareMathOperator*{\argmin}{argmin\,}
\newcommand\norm[1]{\lVert#1\rVert}
\def\x{{\mathbf x}}
\def\z{{\mathbf z}}
\def\bomega{{\boldsymbol{\omega}}}
\def\bepsilon{{\boldsymbol{\epsilon}}}
\def\btheta{{\boldsymbol{\theta}}}
\def\X{{\mathbf X}}
\def\y{{\mathbf y}}
\def\Y{{\mathbf Y}}
\def\Z{{\mathbf Z}}
\def\cN{{\cal N}}
\def\cA{{\cal A}}

\def\I{{\mathbf I}}
\def\R{{\mathbb R}}
\def\C{{\mathbb C}}
\begin{document}

\title{One-Dimensional Deep Image Prior for Curve Fitting of S-Parameters from Electromagnetic Solvers\\
}

\author{
\IEEEauthorblockN{
    Sriram Ravula\IEEEauthorrefmark{1},
    Varun Gorti\IEEEauthorrefmark{1}, 
    Bo Deng\IEEEauthorrefmark{1},
    Swagato Chakraborty\IEEEauthorrefmark{2},
    James Pingenot\IEEEauthorrefmark{2},\\
    Bhyrav Mutnury\IEEEauthorrefmark{3}, 
    Doug Wallace\IEEEauthorrefmark{3},
    Doug Winterberg\IEEEauthorrefmark{3},
    Adam Klivans\IEEEauthorrefmark{1}, and
    Alexandros G. Dimakis\IEEEauthorrefmark{1}
    }
\IEEEauthorblockA{
    \IEEEauthorrefmark{1}University of Texas at Austin,
    \IEEEauthorrefmark{2}Siemens,
    \IEEEauthorrefmark{3}Dell \\
    Email: \{sriram.ravula, vgorti, bodeng\}@utexas.edu, 
           \{swagato.chakraborty, james.pingenot\}@siemens.com, \\
           \{bhyrav.mutnury, doug.wallace, doug.winterberg\}@dell.com,
           klivans@utexas.edu, dimakis@austin.utexas.edu
    }
}

\maketitle

\begin{abstract}
A key problem when modeling signal integrity for passive filters and interconnects in IC packages is the need for multiple S-parameter measurements within a desired frequency band to obtain adequate resolution.  These samples are often computationally expensive to obtain using electromagnetic (EM) field solvers.  Therefore, a common approach is to select a small subset of the necessary samples and use an appropriate fitting mechanism to recreate a densely-sampled broadband representation.  We present the first deep generative model-based approach to fit S-parameters from EM solvers using one-dimensional Deep Image Prior (DIP). DIP is a technique that optimizes the weights of a randomly-initialized convolutional neural network to fit a signal from noisy or under-determined measurements. We design a custom architecture and propose a novel regularization inspired by smoothing splines that penalizes discontinuous jumps.  We experimentally compare DIP to publicly available and proprietary industrial implementations of Vector Fitting (VF), the industry-standard tool for fitting S-parameters. Relative to publicly available implementations of VF, our method shows superior performance on nearly all test examples using only $5-15\%$ of the frequency samples. 
Our method is also competitive to proprietary VF tools and often outperforms them for challenging input instances.
\end{abstract}

\begin{IEEEkeywords}
s-parameter, vector fitting, deep image prior
\end{IEEEkeywords}

\blfootnote{Code available at \url{https://github.com/Sriram-Ravula/Curvefitting-DIP}.}

\section{Introduction}
\label{sec:intro}

\begin{figure*}[!t]%
    \centering
    \subfloat{{\includegraphics[width=0.33\linewidth]{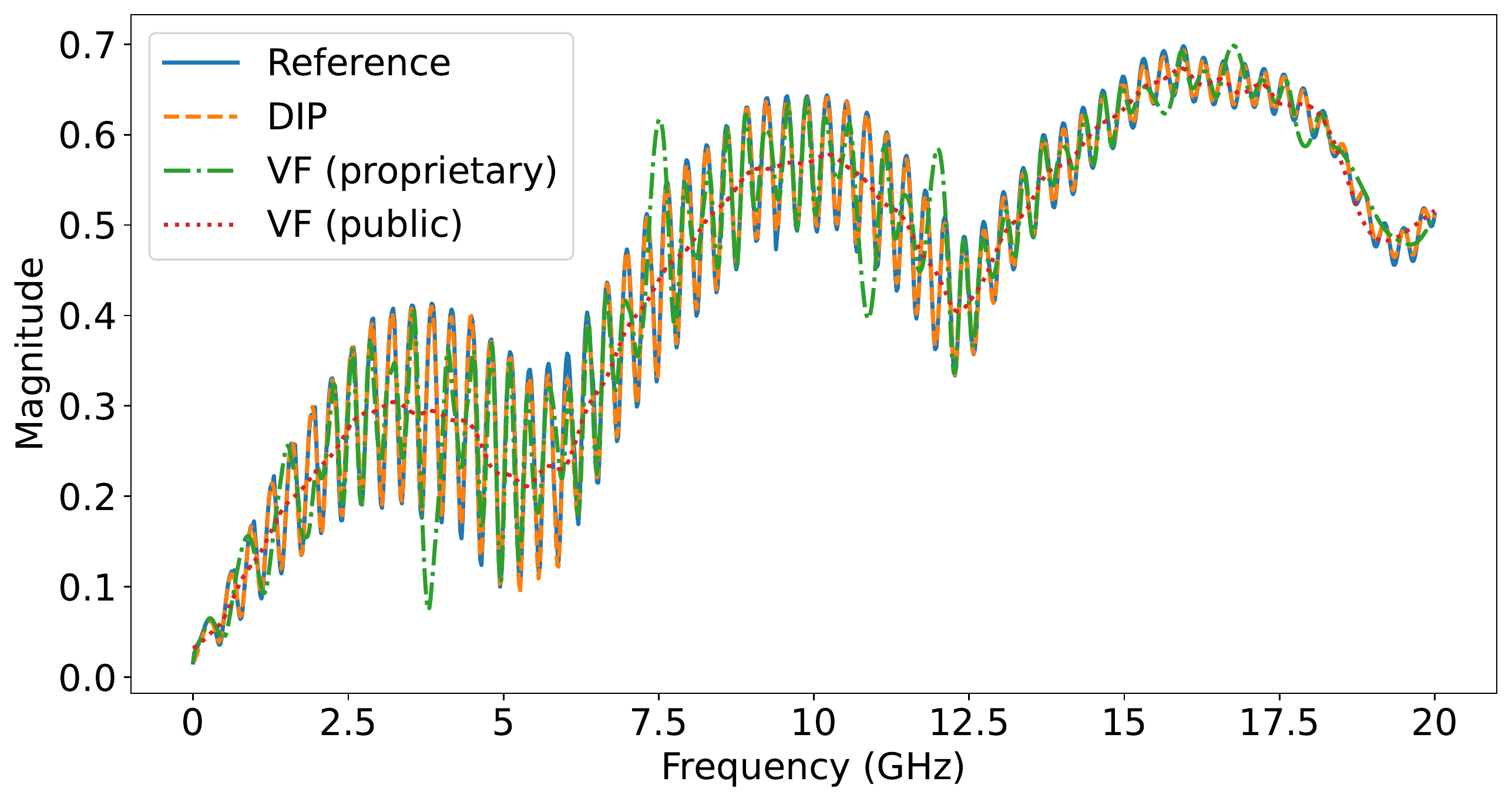} }}
    \subfloat{{\includegraphics[width=0.33\linewidth]{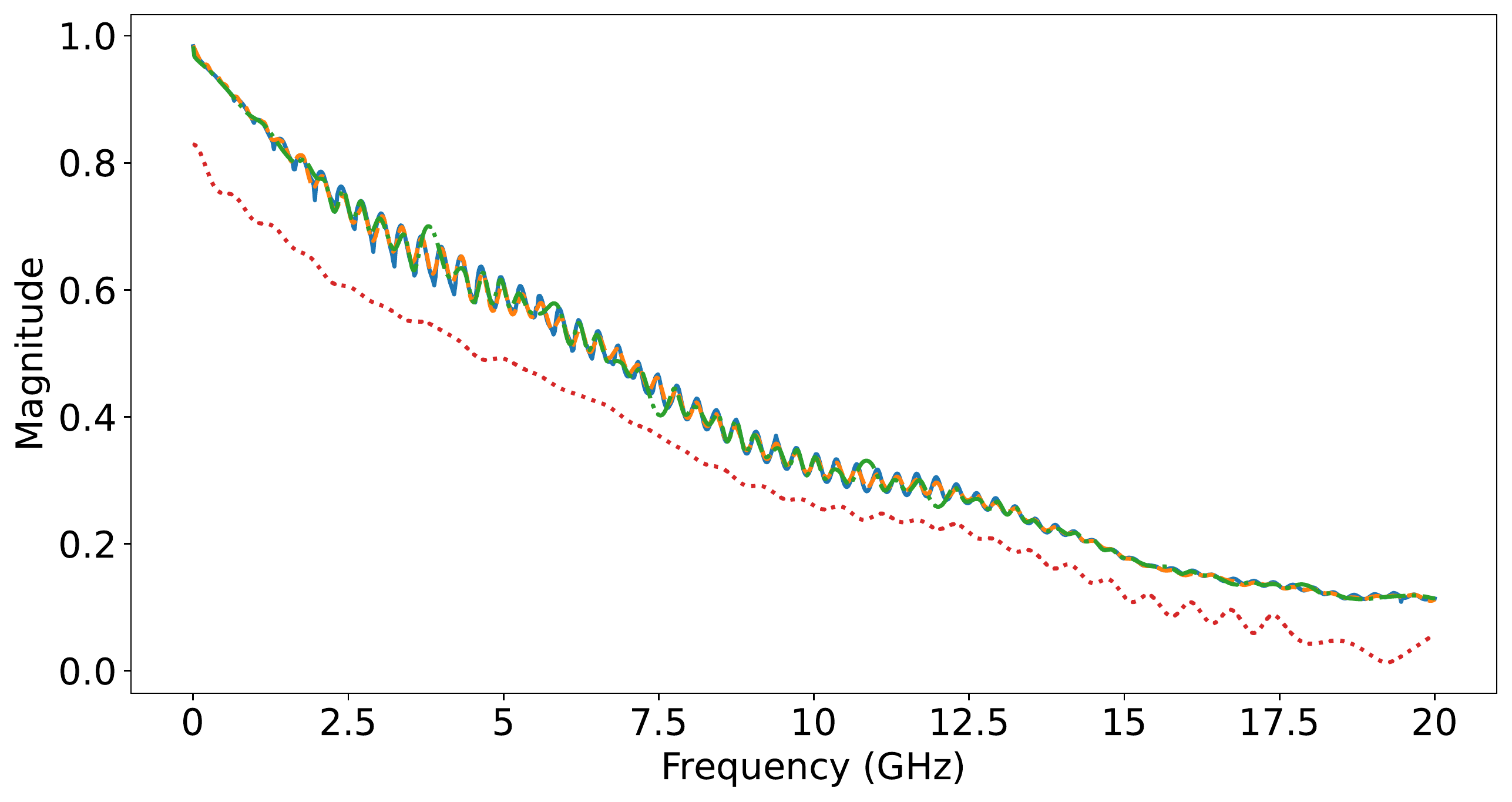} }}%
    \subfloat{{\includegraphics[width=0.33\linewidth]{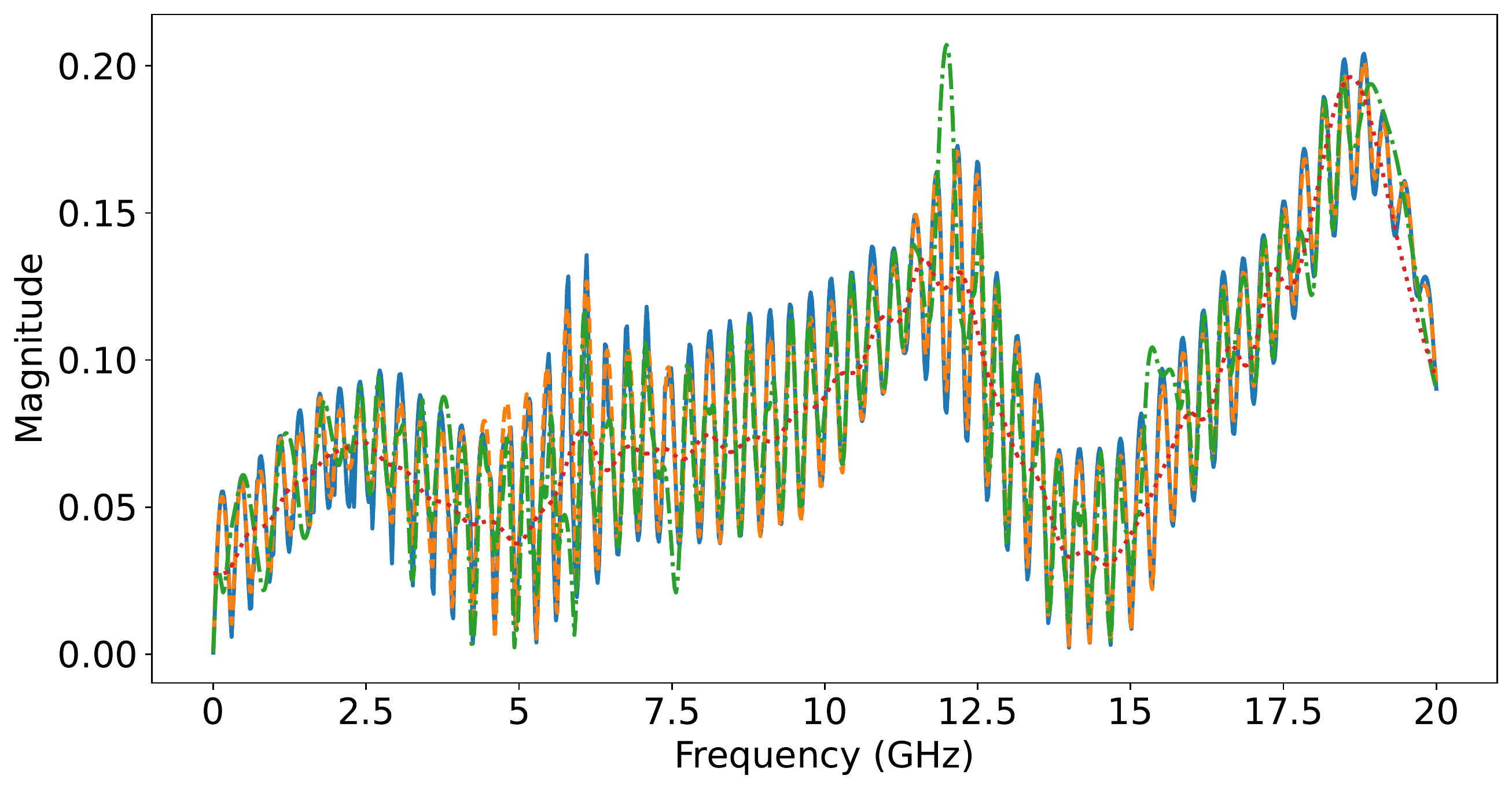} }}
    \vspace{0.01pt}
    \subfloat{{\includegraphics[width=0.33\linewidth]{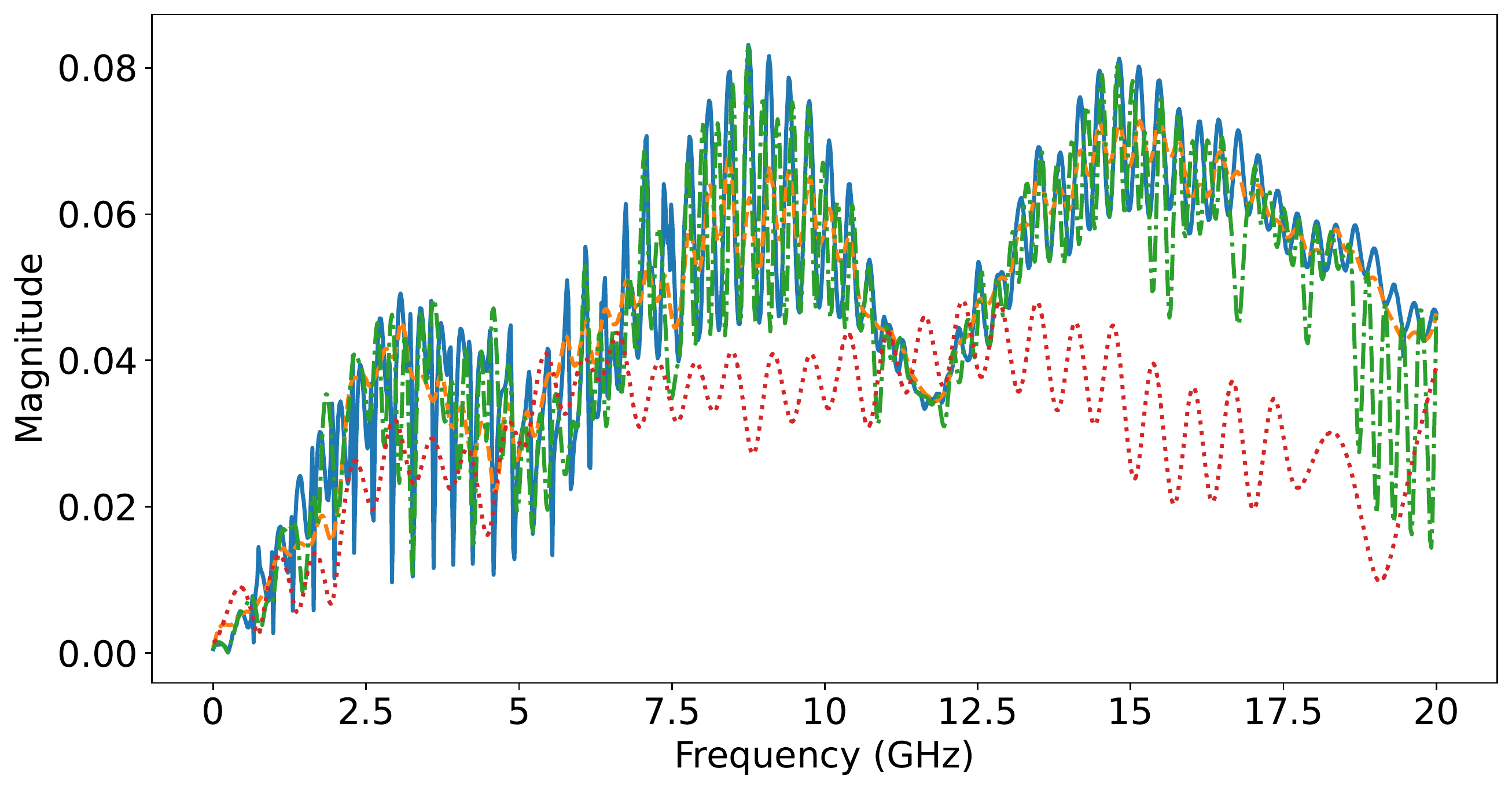} }}%
    \subfloat{{\includegraphics[width=0.33\linewidth]{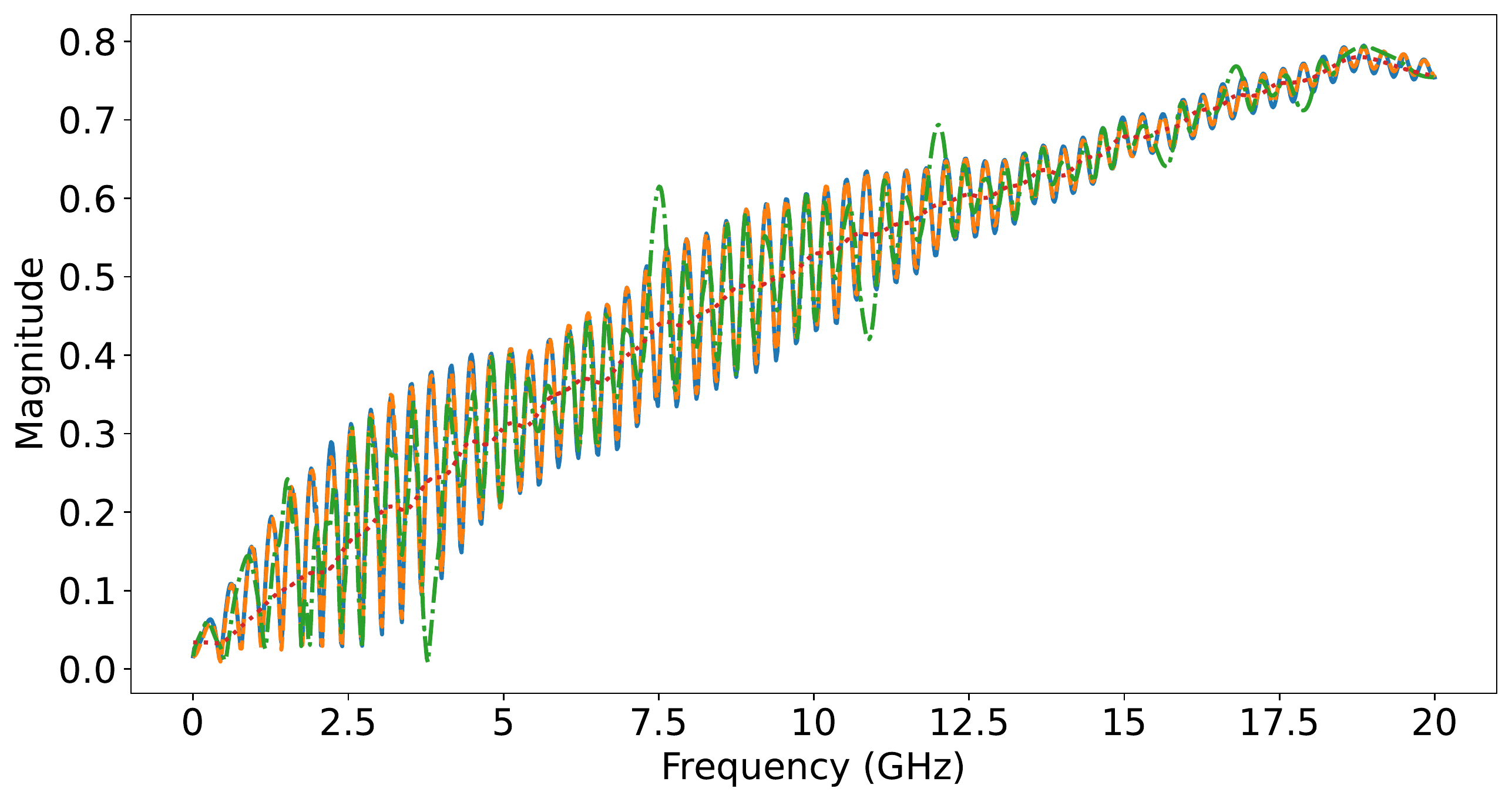} }}%
    \subfloat{{\includegraphics[width=0.33\linewidth]{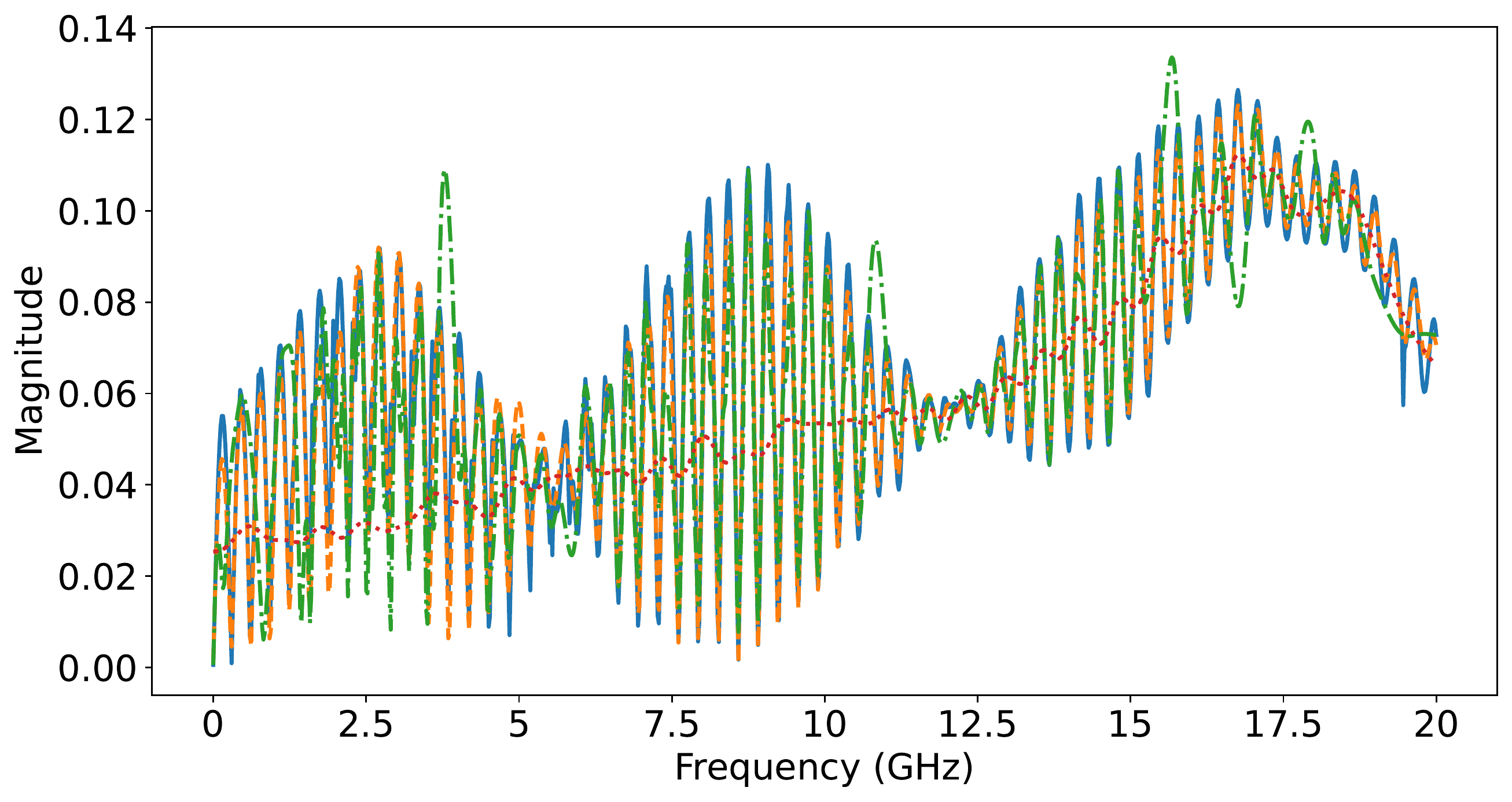} }}%
    \caption{  \textbf{Magnitude Plots for Curve Fitting S-parameters for a 4-port system.} We show magnitude plots for curve fitting a 4-port system using 132 out of a possible 1000 simulated frequency samples. We remove several identical S-parameters before plotting for clarity. The DIP output closely tracks the reference fully-sampled data in each S-parameter while both VF implementations display artifacts in their fitted curves.}
    % \vspace{0.1in}
    \label{fig:qualitative_9}
\end{figure*}

%Demand for high data rate and low power budget necessitate accurate electromagnetic characterization of interconnects in modern electronic systems to ensure acceptable signal integrity. 

Modern electronic system design requires accurate electromagnetic (EM) characterization of interconnects to ensure acceptable signal integrity. 
The interconnects are often designed without solid referencing, due to competitive pricing targets for high volume products and in some cases due to flex interconnects for consumer electronic products with small form factors.  As a result, computationally cheaper EM characterization methods such as transmission line analysis fail to deliver sufficient accuracy at high frequencies and 3D EM simulation is required. Despite recent advances in 3D computational
electromagnetics and parallel solvers~\cite{Fast-solvers,Parallel},
accurate 3D characterization of complex interconnects remains computationally prohibitive. For frequency domain broadband characterization of digital interconnects this expensive computation needs to be carried out at multiple frequencies to provide enough resolution and length of time domain simulation. 

A common practice is to use some fitting techniques where the 3D EM field solver does not run at all of the desired frequency points, and the final frequency sweep output is obtained by fitting a subset of frequencies. Traditionally this interpolation is carried out by model order reduction techniques such as AWE expansion with Pade approximation~\cite{AWE-Pade} or Vector Fitting (VF) \cite{vector_fitting}.  The advantage of the latter is that it allows the core EM field solver to be treated like a black box, and is often preferred for commercial tools, enabling modularity for the frequency domain EM engine. VF is often used in an adaptive loop to predict the next best sampling frequency with a convergence target. Vector Fitting methods represent the incoming samples in a pole-residue form and use a least squares method to obtain the coefficients of the pole-residue expansion.

One limitation of Vector Fitting is its sensitivity to the number of pole pairs used in the expansion. Too few poles result in under-fitting and losing important details of the true frequency domain response, while too many poles  result in over-fitting and non-physical signatures leading to non-passive behavior. In some cases, it is possible to run a parameter search to pick the best number of pole pairs -- however this does not guarantee a relief from over-fitting and is expensive, especially for interconnect models with a large number of ports and complex frequency domain signatures.

We present the first deep generative model-based approach to fit S-parameters from EM solvers. 
We develop a novel interpolation technique using the framework of Deep Image Prior (DIP)~\cite{dip}, a powerful  deep learning method that optimizes the weights of a randomly-initialized convolutional neural network to fit a signal from noisy or under-determined measurements. 

\subsection{Our Contributions}
\label{sec:contributions}
\begin{itemize}
\item We construct the first deep generative prior for S-parameters. Our design uses a custom architecture and a regularization inspired by smoothing splines that penalizes discontinuous jumps.
\item Our method requires no pre-training, and can interpolate curves using only 5\textendash15\% of the required output frequency samples. 
\item We introduce a novel sparsity constraint on third derivatives, resulting in improved approximations in PSNR. 
\item We outperform public Vector Fitting implementations by an order of magnitude, and give improvements to proprietary methods on electrically-long channels (Figure~\ref{fig:quant_skrf} and Table~\ref{tab:results}).
\item Our method can produce multiple samples from the posterior distribution, allowing us to obtain uncertainty estimates that are well-correlated with the real error (Figure~\ref{fig:uncertainty}).  
\end{itemize}

\begin{table*}[!t]
\centering
\caption{\textbf{Specifications for the Examples in our Dataset.} We list the number of ports ($p$), total number of frequency points ($f$), and the percentage of frequencies observed by the final proprietary VF output using active learning, for each of the examples in our simulated dataset. }
\label{tab:specs}
\begin{tabular}{@{}rccccccccccccccc@{}}
\toprule
Index       & 0    & 1   & 2    & 3   & 4    & 5    & 6    & 7    & 8    & 9    & 10   & 11   & 12   & 13   & 14   \\ \midrule
Ports       & 4    & 2   & 2    & 2   & 2    & 16   & 12   & 12   & 12   & 16   & 12   & 16   & 16   & 12   & 12   \\
Frequencies & 1000 & 341 & 1500 & 600 & 2081 & 7000 & 9999 & 9999 & 2000 & 9999 & 9000 & 8000 & 2000 & 3000 & 3000 \\
\begin{tabular}[c]{@{}r@{}}\% Observed\end{tabular} &
  13.2 &
  6.74 &
  4.13 &
  1.83 &
  7.44 &
  0.38 &
  2.01 &
  2.01 &
  1.00 &
  0.67 &
  0.48 &
  0.48 &
  5.85 &
  2.83 &
  0.86 \\ \bottomrule
\end{tabular}
\end{table*}

\begin{table*}[!t]
\centering
\caption{\textbf{Peak Signal-to-Noise Ratio (PSNR) Results.} We compare DIP and a public implementation of VF to a proprietary implementation of VF for curve fitting each example in our dataset at various sub-sampling rates (the exact rate for each example is given in Table~\ref{tab:specs}). The top row lists the index of each example in our dataset. The best PSNR result for each example is bolded.}
\label{tab:results}
\begin{tabular}{@{}rccccccccccccccc@{}}
\toprule
Index       & 0             & 1    & 2    & 3             & 4    & 5             & 6    & 7    & 8             & 9    & 10   & 11   & 12   & 13   & 14   \\ \midrule
DIP         & \textbf{49.4} & 30.9 & 41.4 & \textbf{12.7} & 39.9 & \textbf{31.7} & 33.3 & 28.1 & \textbf{44.1} & 30.1 & 25.4 & 35.4 & 51.5 & 48.1 & 44.0 \\
VF (proprietary) &
  37.3 &
  \textbf{58.5} &
  \textbf{43.8} &
  11.4 &
  \textbf{42.2} &
  31.1 &
  \textbf{40.2} &
  \textbf{29.2} &
  38.3 &
  \textbf{30.2} &
  \textbf{49.5} &
  \textbf{42.3} &
  \textbf{53.2} &
  \textbf{50.0} &
  \textbf{59.4} \\
VF (public) & 32.4          & 29.3 & 34.2 & 11.2          & 16.6 & 27.0          & 35.0 & 28.6 & 29.1          & 27.5 & 35.4 & 39.1 & 31.7 & 30.7 & 39.8 \\ \bottomrule
\end{tabular}
\end{table*}

\section{Background}
\label{sec:background}

\subsection{Curve Fitting S-parameters}
\label{sec:curve_fitting}

In our problem setting, there is a generating function $f(\cdot): \R \rightarrow \C^{p \times p}$ that takes frequency values as input and produces complex-valued matrices as output. Here, $p$ is the number of ports in the EM system. For a given frequency $\omega$, we denote the resulting matrix as $f(\omega) \coloneqq \X_{\omega}$, with the individual entries $\X_{\omega}[i, j] \in \C, \:\: i, j \in \{0, \dots, p-1\}$ termed $\emph{S-parameters}$. %For the special case of \emph{reciprocal} systems, the matrices $\X_{\omega}$ are symmetric $\forall \omega$. 

It is generally impossible to create an analytical model of a channel's frequency response. Fortunately, for most applications it suffices to have a set of densely-sampled points from the underlying function. For a given vector of query frequencies $\bomega = [\omega_0, \dots, \omega_{f-1}]$, we wish to obtain the samples $\{\X_{\omega_0}, \dots, \X_{\omega_{f-1}}\}$. We denote by $\bar{\X} \in \C^{p \times p \times f}$ the concatenation of the sample matrices along a new frequency dimension, where $\bar{\X}[i, j, k] = \X_{\omega_{k}}[i, j]$, for $i, j \in \{0, \dots, p-1\}$ and $k \in \{0, \dots, f-1\}$. Therefore, our ultimate goal is to find $\bar{\X}$. 

Given the physical specifications of a system and a query frequency $\omega$, an EM field solver performs simulations to output $\X_{\omega}$ at the requested frequency. However, field solvers are extremely computationally expensive, with simulations for the most complex systems taking up to several hours per frequency point on modern, high-performance workstations with a large number of compute cores.

The computational requirement is typically reduced by querying a vector of sub-sampled frequencies $\bomega_{y} = [\omega_{y_0}, \dots, \omega_{y_{f'- 1}}]$, where the new index set is a subset of the original query index set: $\{{y_0}, \dots, {y_{f'- 1}}\} \subseteq \{0, \dots, f-1\}$, and $f' \leq f$ is the number of sub-sampled frequency points. We again concatenate the resultant sub-sampled outputs $\{\X_{\omega_{y_0}}, \dots, \X_{\omega_{y_{f'- 1}}}\}$ along a new frequency dimension and call the resulting complex-valued tensor $\bar{\Y} \in \C^{p \times p \times f'}$ the \emph{measurements} or \emph{observations}. The goal of curve-fitting is to recover the densely-sampled frequency series $\bar{\X}$ given the sub-sampled measurements $\bar{\Y}$.

\subsection{Vector Fitting}
\label{sec:vector_fitting}

The most widely-used method for curve fitting S-parameters is Vector Fitting \cite{vector_fitting}, which directly parameterizes a functional approximation of the underlying generating function $f(\cdot)$. This method relies on the observation that the frequency response of a linear system can be modeled as a complex-valued rational function of the form
\begin{equation}
\label{eq:vector_fit}
    h(s) = \mathbf{d} + s\mathbf{e} + \sum_{k=0}^{K-1} \frac{\mathbf{c}_k}{s - p_k},
\end{equation}
where the functional model is $h(\cdot): \C \rightarrow \C^{p \times p}$, the argument $s = \sigma + j\omega$ is defined in the Laplace domain, and the goal is that the fitted model matches the measurements at every observed frequency point: $h(s = j\omega_t) = \X_{\omega_t},\:\: \forall t \in \{{y_0}, \dots, {y_{f'- 1}}\}$. Here, $\mathbf{d} \in \R^{p \times p}$ is the DC bias at $\omega = 0$, $\mathbf{e} \in \R^{p \times p}$ is a scale factor for frequency-dependent bias, $\mathbf{c} \in \C^{K \times p \times p}$ is a tensor of complex residues with matrix-valued entries $\mathbf{c}_k \in \C^{p \times p}$ for $k \in \{0, \dots\, K-1\}$, and $\mathbf{p} \in \C^K$ is a vector of complex poles with entries $p_k \in \C$. The number of poles, $K$, is a hyperparameter chosen using heuristics or cross-validation. The parameters $\mathbf{d}, \mathbf{e}, \mathbf{c},$ and $\mathbf{p}$ are optimized iteratively, either for a predetermined number of steps or until the model fits the measurements to a specific tolerance. Subsequent additions to the original Vector Fitting method have included improvements to the iterative algorithm \cite{vector_fit_pole_relocation} and enforcement of physical constraints on the underlying system \cite{vector_fit_passivity}. While fast relative to 3D field solvers, VF operations with a large number of poles can be computationally expensive for systems with high numbers of ports and frequency samples \cite{vf_complexity}.

\subsection{Deep Image Prior}
\label{sec:dip}

Deep Image Prior is an unsupervised learning framework that uses the convolutional architecture of an untrained neural network as a prior for solving image inverse problems \cite{dip}. Suppose we are given $\y = \cA(\x^*) + \bepsilon$, where $\y \in \R^m$ are noisy linear measurements, $\cA(\cdot): \R^{m} \rightarrow \R^n$ is a known, differentiable forward operator, $\x^* \in \R^{n}$ is an unknown image, and $\bepsilon \in \R^{m}$ is additive white noise. Usually the system is underdetermined, i.e. $m < n$, and has infinitely many possible solutions. Because of this, we need some form of regularization or \emph{prior} to constrain the set of solutions. 

DIP finds a solution $\hat{\x}$, defined as
\begin{equation}
\label{eq:dip}
    \hat{\x} = G_{\btheta^*}(\z), \:\:\: \btheta^* = \argmin_{\btheta} \norm{\y - \cA(G_{\btheta}(\z))}_2^2,
\end{equation}
where $G_{\theta}$ is an \emph{untrained} convolutional neural network (cnn), parameterized by \emph{randomly-initialized} weights $\btheta \in \R^{d}$, with latent input vector $\z \in \R^{k}$ (which can be optimized along with $\btheta$). Even though the network has not been trained and only gets to see noisy measurements of a single image, DIP achieves high-quality reconstructions for various problems such as denoising, super-resolution, and inpainting \cite{dip}. The central idea is that the deep convolutional generator structure allows it to fit natural signals faster than noise, which explains its ability to denoise and interpolate.       

Numerous works have developed the DIP framework in various directions. The authors of One-dimensional DIP demonstrate that deep networks with one-dimensional convolutions are a good prior for time series inverse problems \cite{1d_dip}. Other works take a Bayesian viewpoint of DIP for uncertainty quantification \cite{dip_uncertainty_tomography, dip_uncertainty_dropout, dip_uncertainty_mean_field}. Bayes' Rule gives the relationship between the \emph{posterior} $p(\btheta|\y)$ that we want to characterize, \emph{likelihood} $p(\y|\btheta)$ we get from the measurements, and prior $p(\btheta)$ imposed by our model as $p(\btheta|\y) \propto p(\y|\btheta) p(\btheta)$. The authors of \cite{dip_bayesian} analyze DIP from a Bayesian perspective and use Stochastic Gradient Langevin Dynamics (SGLD) as an alternative to gradient descent for updating DIP weights. By introducing a regularization term to the objective in Eq.~\eqref{eq:dip} and adding noise to the weights after each update step, SGLD performs posterior sampling (assuming some other suitable conditions are met). The SGLD update for one iteration is given by
\begin{equation}
\begin{aligned}
\label{eq:sgld}
    \Delta_{\btheta} &= \frac{\alpha}{2} \bigg[ \nabla_{\btheta} \log p(\y | \btheta) + \nabla_{\btheta} \log p(\btheta) \bigg] + \mathbf{n}, \\ 
    \mathbf{n} &\sim \cN(\mathbf{0}, \alpha \I),
\end{aligned}
\end{equation}
where $\alpha$ is the step size, $\nabla_{\btheta} \log p(\y | \btheta)$ is the gradient of the measurement loss in Eq.~\eqref{eq:dip} (assuming Gaussian noise and with additional scale factors), and $\nabla_{\btheta} \log p(\btheta)$ is the gradient of a regularization function such as $l_2$ weight decay. SGLD eliminates the need for early stopping during DIP optimization, which was used to prevent the network from overfitting to the measurements. In addition, SGLD has the benefit of allowing us to sample multiple reconstructions $\hat{\x}_i = G_{\btheta_i}(\z), \:\:\: \btheta_i \sim p(\btheta|\y)$ instead of producing a point estimate.  

\begin{figure}[!t]
    \centering
    \includegraphics[width=0.7\linewidth]{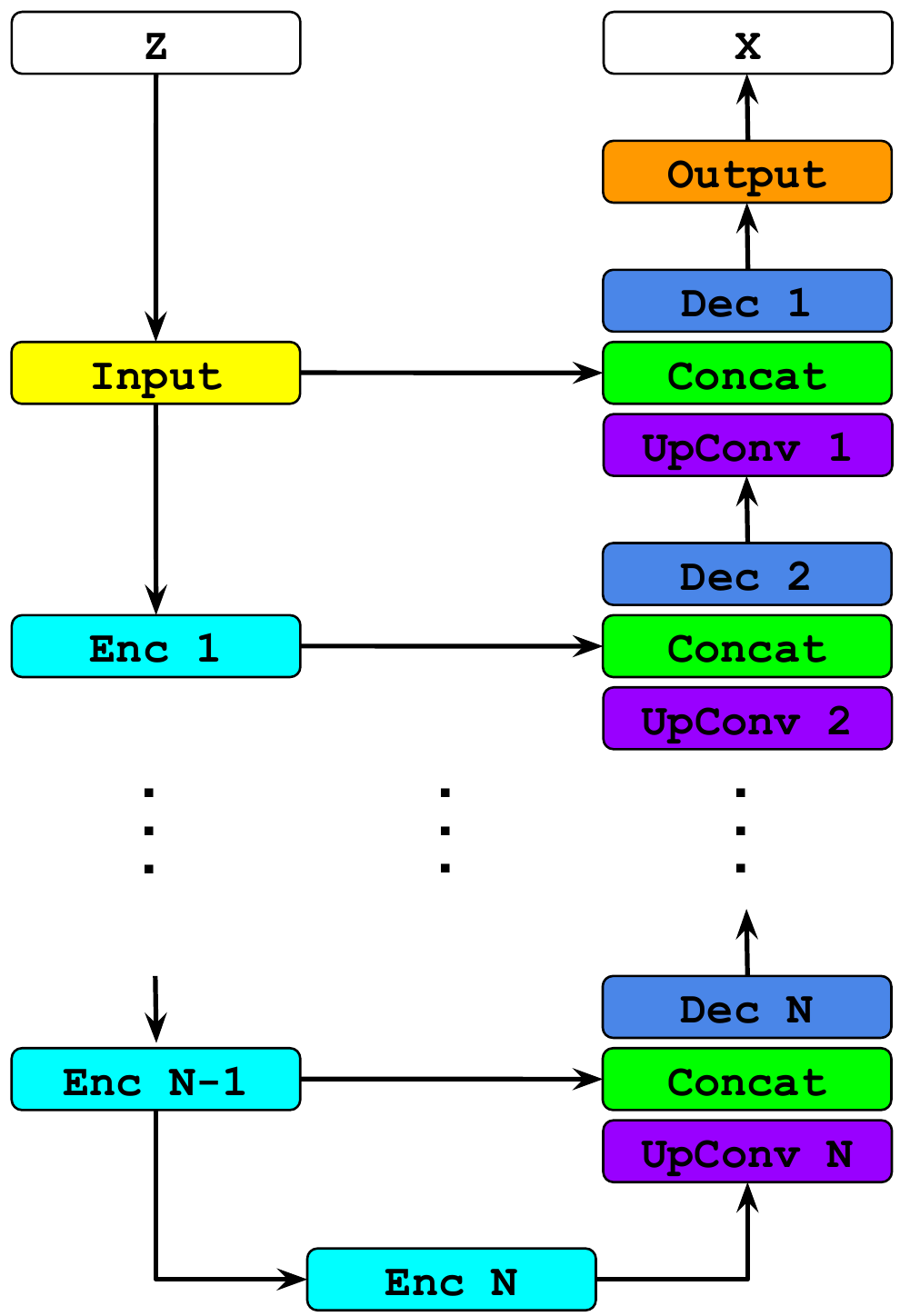}
    \caption{\textbf{DIP Network Architecture.} Our network is based on that of One-dimensional DIP \cite{1d_dip}, which uses a U-Net style architecture \cite{ronneberger2015u}. The index $i$ denotes the $i$-th layer of a network, with $i \in \{1,\dots, N\}$ for a network with depth $N$. The ``Concat'' block denotes a concatenation operation performed along the channel dimension.}
    \label{fig:Arch}
\end{figure}

\begin{figure}[!t]
    \centering
    \includegraphics[width=0.7\linewidth]{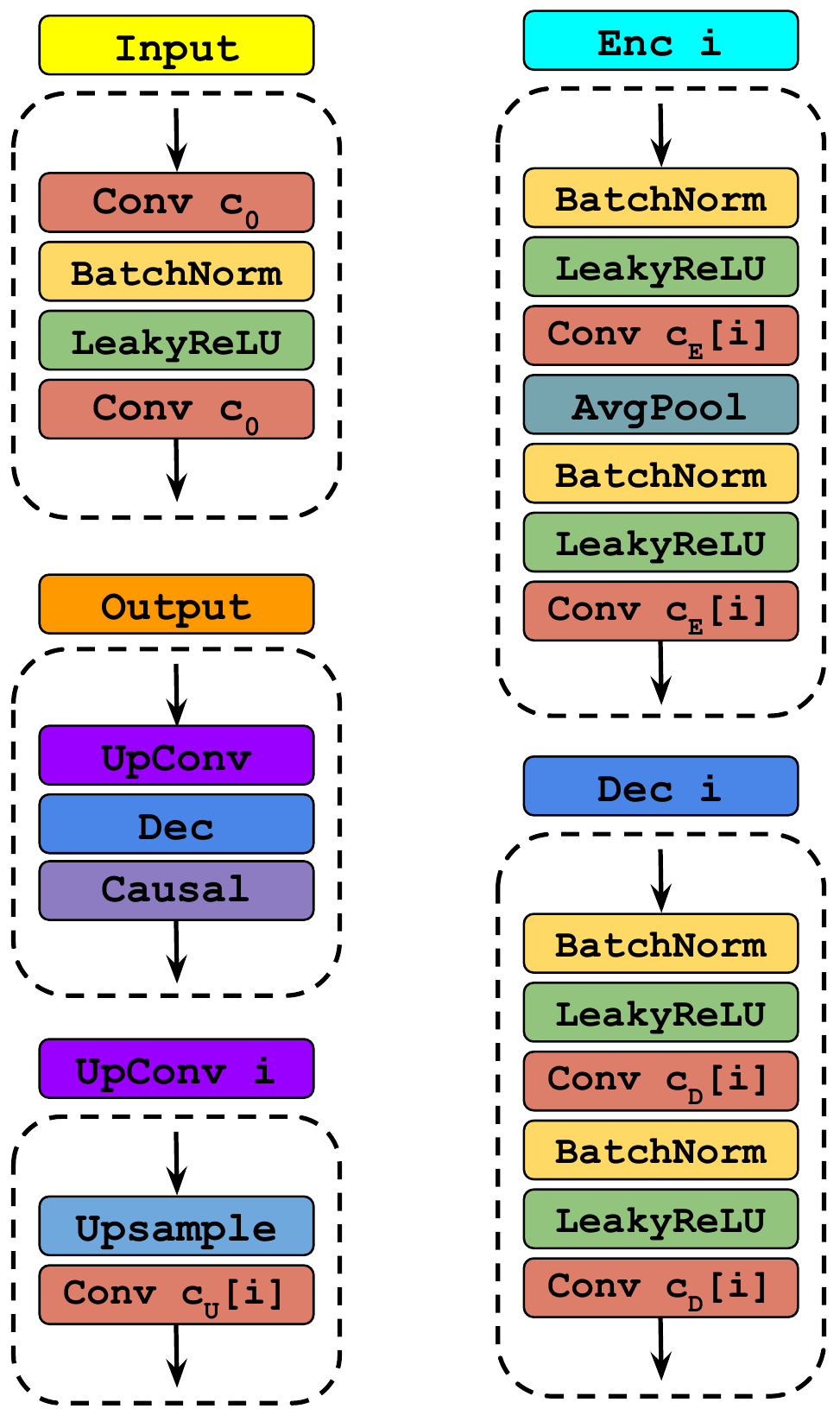}
    \caption{\textbf{Breakdown of Network Blocks.} We list the operations performed by each component of the network. Here, ``Enc'' is an encoder layer, ``Dec'' is a decoder layer, and ``UpConv'' is an upsampling layer. The components are indexed by $i$, which denotes the $i$-th layer of a network. The convolutions are parameterized by values $c_0, c_E[i], c_D[i],$ and $c_U[i]$, which correspond to the number of filters in the input, encoder, decoder, and upsampling layers, respectively. }
    \label{fig:components}
\end{figure}

\subsection{Enforcing Causality}
\label{sec:causal_passive}

S-parameters from linear time-invariant (LTI) systems are causal. For a time-domain signal $h(t)$, causality is defined as the property that the signal cannot produce a response before being given an input, otherwise defined as
\begin{equation}
\label{eq:causality}
    h(t) = 0, \:\:\:\:\:\: \forall t < 0.
\end{equation}
An equivalent frequency-domain definition is given by the Kramers-Kronig relations \cite{kkr} which state that the real and imaginary portions of the signal $H(\omega) = \mathcal{F}(h(t))$, where $\mathcal{F}$ is the Fourier transform, are related by the Hilbert transform $\mathcal{H}$ as
\begin{equation}
\label{eq:kkr}
    \textup{Im}(H(\omega)) = - \mathcal{H}(\textup{Re}(H(\omega))), \:\:\:\:\:\: \forall \omega.
\end{equation}
% Passivity is the condition that a system cannot add energy to an input. A matrix-valued multi-port system $f(\omega) = \X_{\omega}$ (as in Section~\ref{sec:curve_fitting}) is considered passive if it satisfies
% \begin{equation}
% \label{eq:passive}
%     \sigma_{max}(\omega) \leq 1, \:\:\:\:\:\: \forall \omega \in \Omega,
% \end{equation}
% where $\sigma_{max}(\omega)$ is the largest singular value of $\X_{\omega}$ and $\Omega$ is the band of interest. In the case of discrete frequency samples, we replace $\Omega$ with the vector of quesry frequencies $\bomega$. 

Enforcing causality is not as simple as predicting the imaginary part from the real part (or vice-versa) using Eq.~\eqref{eq:kkr} due to errors in the simulated data caused by truncation and discretization. Truncation error results from the fact that the Hilbert transform in Eq.~\eqref{eq:kkr} is calculated as an integral over frequency values from $-\infty$ to $\infty$, while the simulated data available to us are band-limited. Discretization error arises from the quantization of the frequency response to fit in floating-point representation.

Torun et. al propose a causality enforcement layer (CEL) at the output of a neural network to mitigate truncation and discretization errors while enforcing causal outputs \cite{causal_passive}. The CEL is parameter-free and applies a series of transformations to a real-only input to create the complex output. The input to the CEL is the real part of the reconstructed frequency response, extrapolated by a factor $n_e$ to reduce truncation error. To create a symmetrical curve around DC, the initial extrapolated response is mirrored, where the positive frequency values are reflected onto the negative frequency spectrum. The Fourier transform of the mirrored spectrum is taken, and the non-DC frequencies are scaled by a factor of 2 before the signal is zero-padded to increase its length by a factor of $n_k$. This padding serves as an interpolation in the frequency spectrum and reduces discretization error. The Inverse Fourier transform of the scaled and padded signal is taken and the resulting signal is truncated to an interpolation factor of $n_k$ (down from the total $n_e \cdot n_k$ factor). The final output is a complex signal whose real and imaginary portions are those of the truncated, transformed signal, scaled by $n_k$ and $-n_k$ respectively. The CEL is applied concurrently on all channels of the input, treating them as independent frequency-domain signals.      

\begin{figure*}[!t]%
    \centering
    \subfloat{{\includegraphics[width=0.49\linewidth]{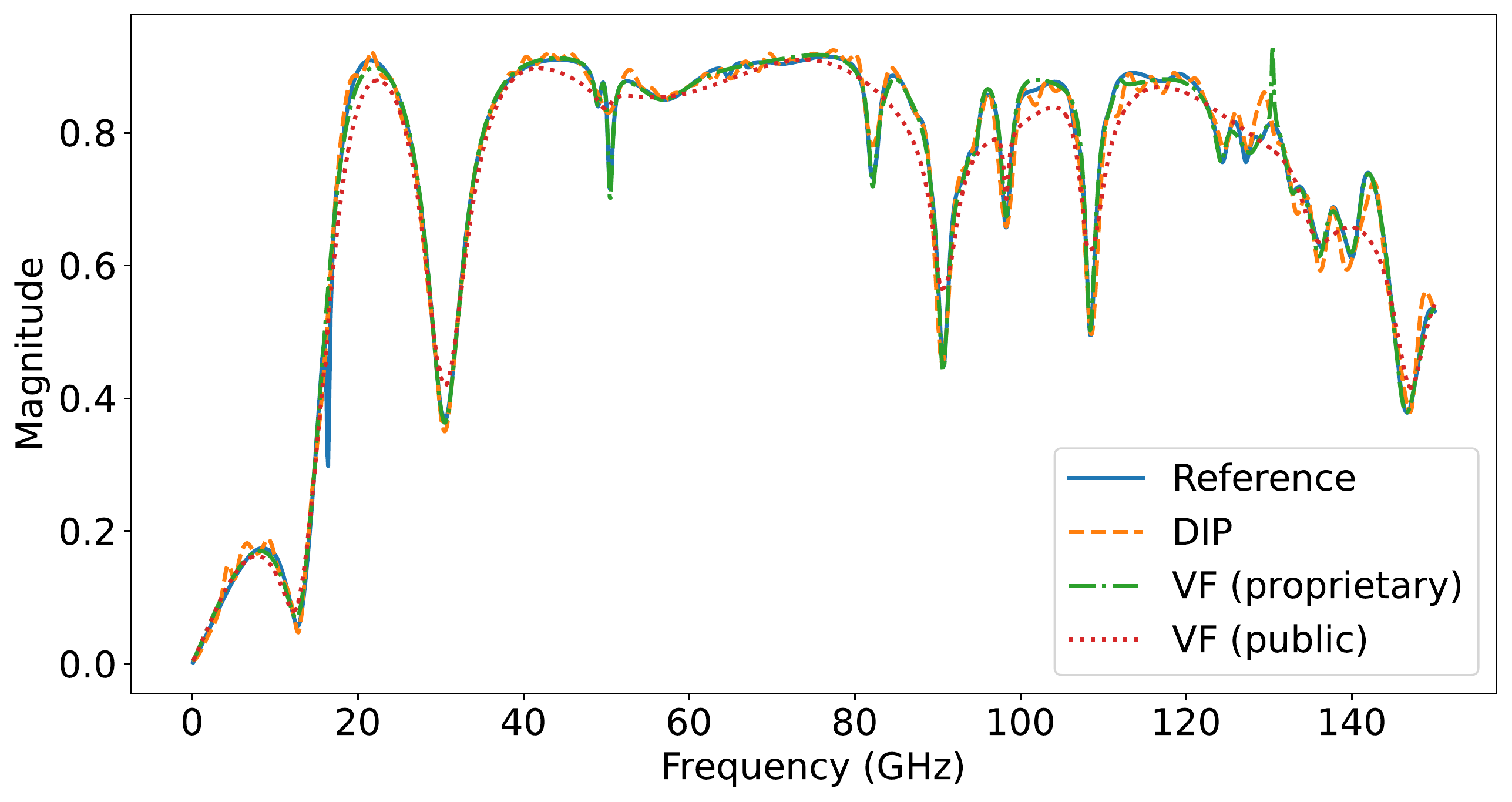} }}
    \subfloat{{\includegraphics[width=0.49\linewidth]{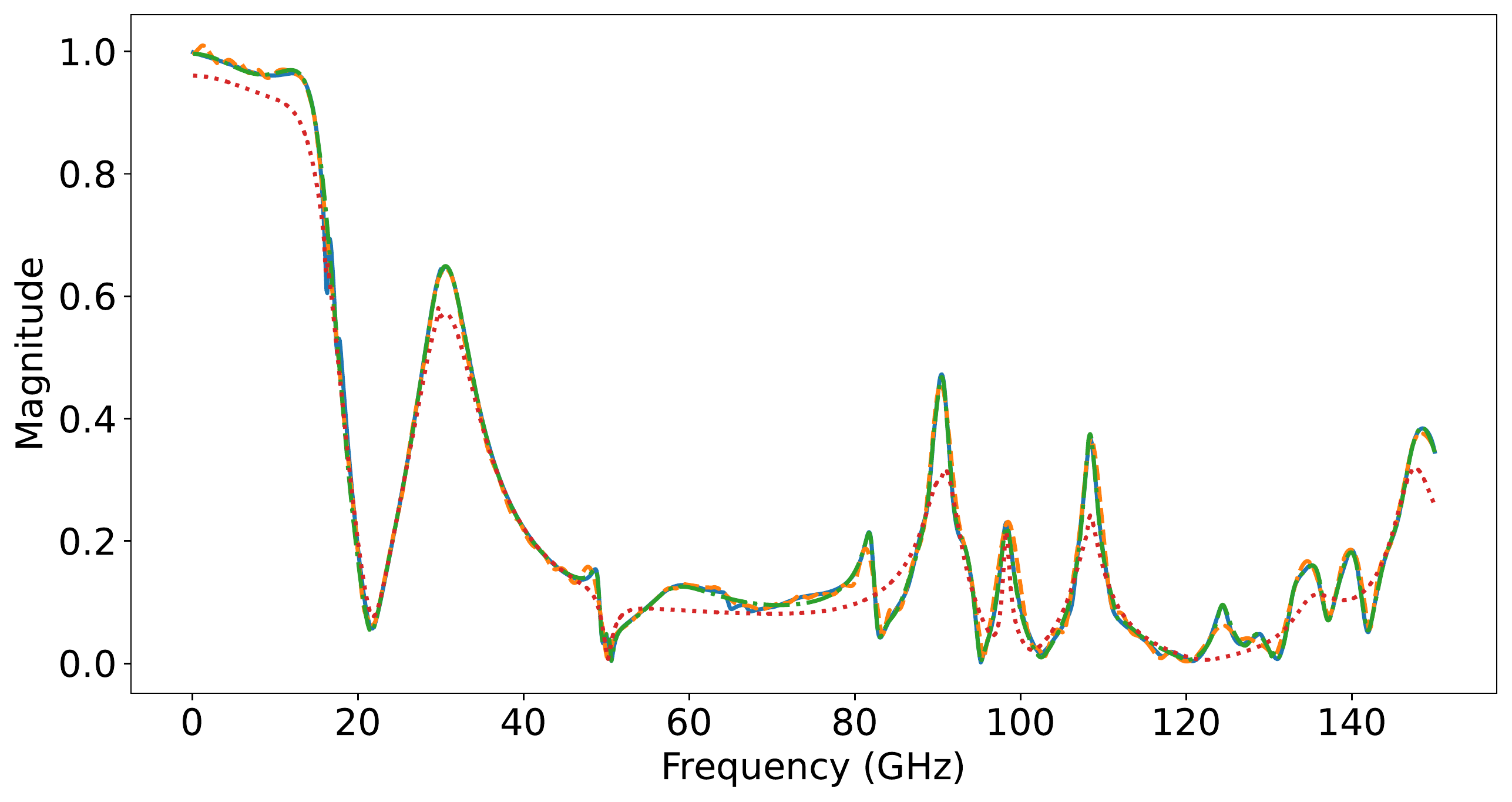} }}%
    \caption{  \textbf{Magnitude Plots for Curve Fitting S-parameters for a 2-port system.} We show magnitude plots for curve fitting a 2-port system using 62 out of a possible 1500 simulated frequency samples. We remove several identical S-parameters before plotting for clarity. Both the DIP and proprietary VF outputs closely track the reference fully-sampled data while the public VF implementation displays artifacts in its fitted curves.}
    % \vspace{0.1in}
    \label{fig:qualitative_20}
\end{figure*}

\section{Method}
\label{sec:method}

We pose the problem of curve fitting sub-sampled S-parameter measurements as a linear inverse problem and propose to solve it using a novel method based on Deep Image Prior. We consider measurements $\Y = \cA(\X) + \mathcal{E}$, where $\Y \in \R^{r \times f'}$ and $\X \in \R^{r \times f}$ are flattened, real-valued versions of $\bar{\Y}$ and $\bar{\X}$, $\cA(\cdot): \R^{r \times f} \rightarrow \R^{r \times f'}$ is a sub-sampling operator that discards the rows of its argument except for those corresponding to frequencies present in the measurements, and $\mathcal{E} \in \R^{r \times f'}$ is additive white noise. The value $r = 2 p^2$ represents the flattening of each of the $p^2$ entries of the matrices $\X_{\omega}$ into a single dimension, with the factor of 2 due to converting the real and imaginary components into 2 real values. This type of problem is known as \emph{inpainting} for images and \emph{imputation} for time series.  %For the special case of reciprocal systems, we have $r = 2(p(p+1)/2) = p^2 + p$ due to the symmetry. 

We consider the signals $\Y$ and $\X$ multi-variate or multi-channel frequency series. We parameterize a one-dimensional convolutional generator network $G_{\theta}(\cdot): \R^{r \times f} \rightarrow \R^{r \times f}$ and pose our initial problem as:
\begin{equation}
\label{eq:our_method}
    \hat{\X} = G_{\btheta^*}(\Z), \:\:\: \btheta^* = \argmin_{\btheta} \norm{\Y - \cA(G_{\btheta}(\Z))}_2^2,
\end{equation}
where $\Z \in \R^{r \times f}$ is a latent input vector. In the proceeding sections, we first detail the architecture of our generator and how the problem domain informs our choices. Inspired by smoothing splines, we also introduce a regularization term to the problem in Eq.~\eqref{eq:our_method} that encourages a form of smoothness in the network output. We finally describe our optimization algorithm, SGLD, along with our choice of $\Z$.  

\subsection{Architecture}
\label{sec:arch}

The choice of architecture is crucial to the success of DIP, as the method relies on the implicit prior provided by the network structure. We must be careful to pick an architecture that is sufficiently expressive to model realistic S-parameters while also providing good enough regularization to reject noise. We use the base architecture from One-dimensional DIP \cite{1d_dip}, which is based on the U-Net family of autoencoders \cite{ronneberger2015u}. We show a block diagram of the architecture in Figure~\ref{fig:Arch} and each network component in Figure~\ref{fig:components}.   

The network is comprised of an encoder pathway with successively smaller frequency resolution in each layer, followed by a decoder pathway with increasing spatial frequency resolution. Encoder and decoder layers with the same frequency resolution are linked by skip connections, which concatenate the intermediate representations along the channel dimension. As reported by Ulyanov et. al for the case of images, the skip connections aid our network in generating multi-channel S-parameter signals with features from varying frequency scales \cite{dip}. We fix the size of the convolutions to $3$, which, combined with upsampling and downsampling operations, we find to provide sufficient coupling between adjacent frequency values. 

We use batch normalization, LeakyReLU nonlinearity, average pooling for downsampling, and linear interpolation for upsampling as in the original One-dimensional DIP implementation. We alter the final layer to add a causality enforcement layer at the output. The CEL is preceeded by an upsampling layer that increases the frequency resolution of the second-to-last-layer from $f$ frequency points to $2f$ points to meet the extrapolation requirement of the CEL (and therefore we set the extrapolation factor $n_e = 2$). This is followed by a decoder layer that reduces the channel width of the intermediate representation from $c_D[1]$ to $r/2$ to isolate only the real components of the inputs to the CEL. We set the interpolation factor $n_k = 1$, meaning that we do not interpolate the signal. 

\subsection{Smoothing Regularization}
\label{sec:regularization}

\begin{figure*}[!t]%
    \centering
    \subfloat[Fully-sampled Reference]{{\includegraphics[width=0.49\linewidth]{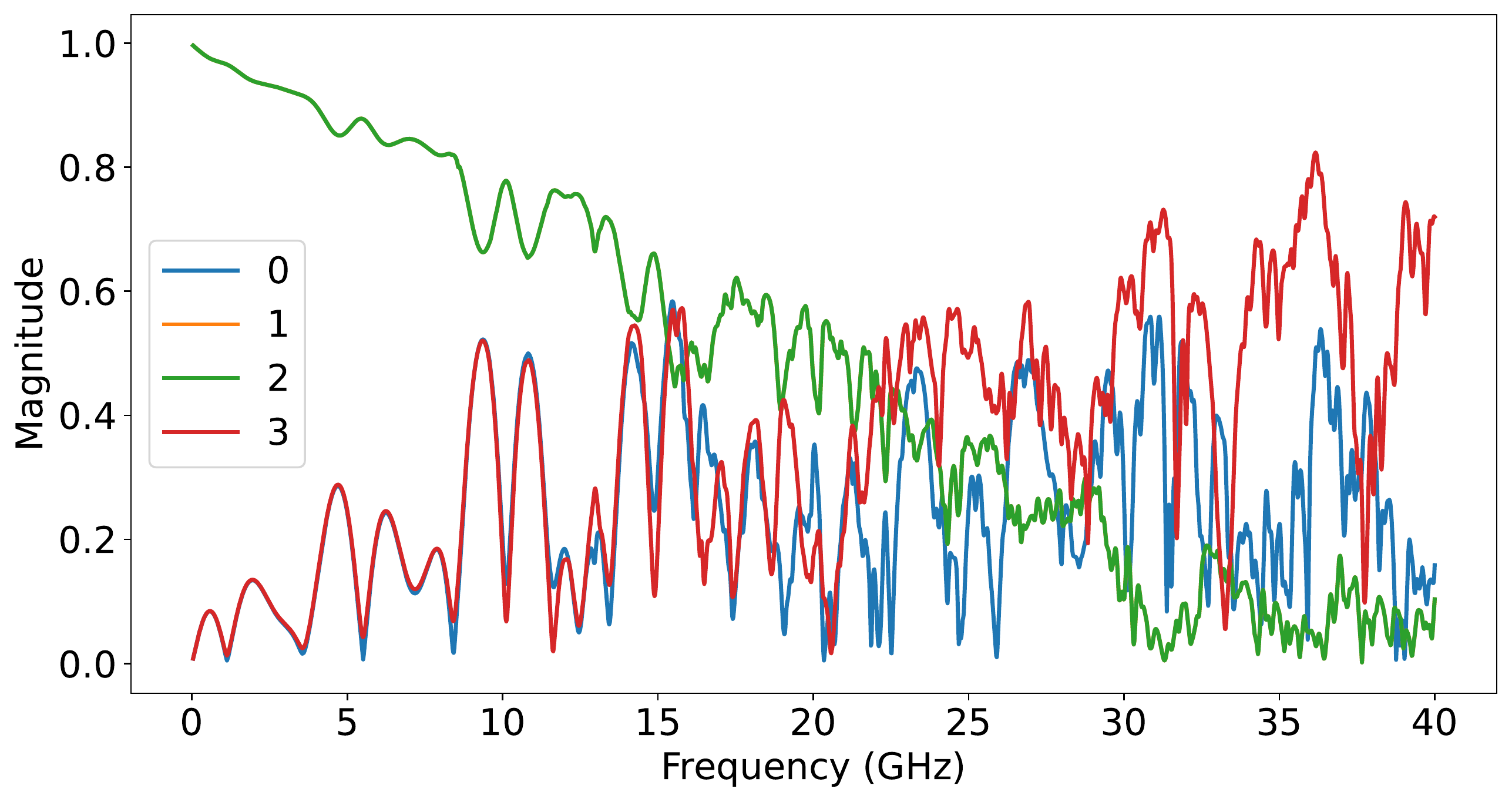} }}
    \subfloat[Mean DIP Output]{{\includegraphics[width=0.49\linewidth]{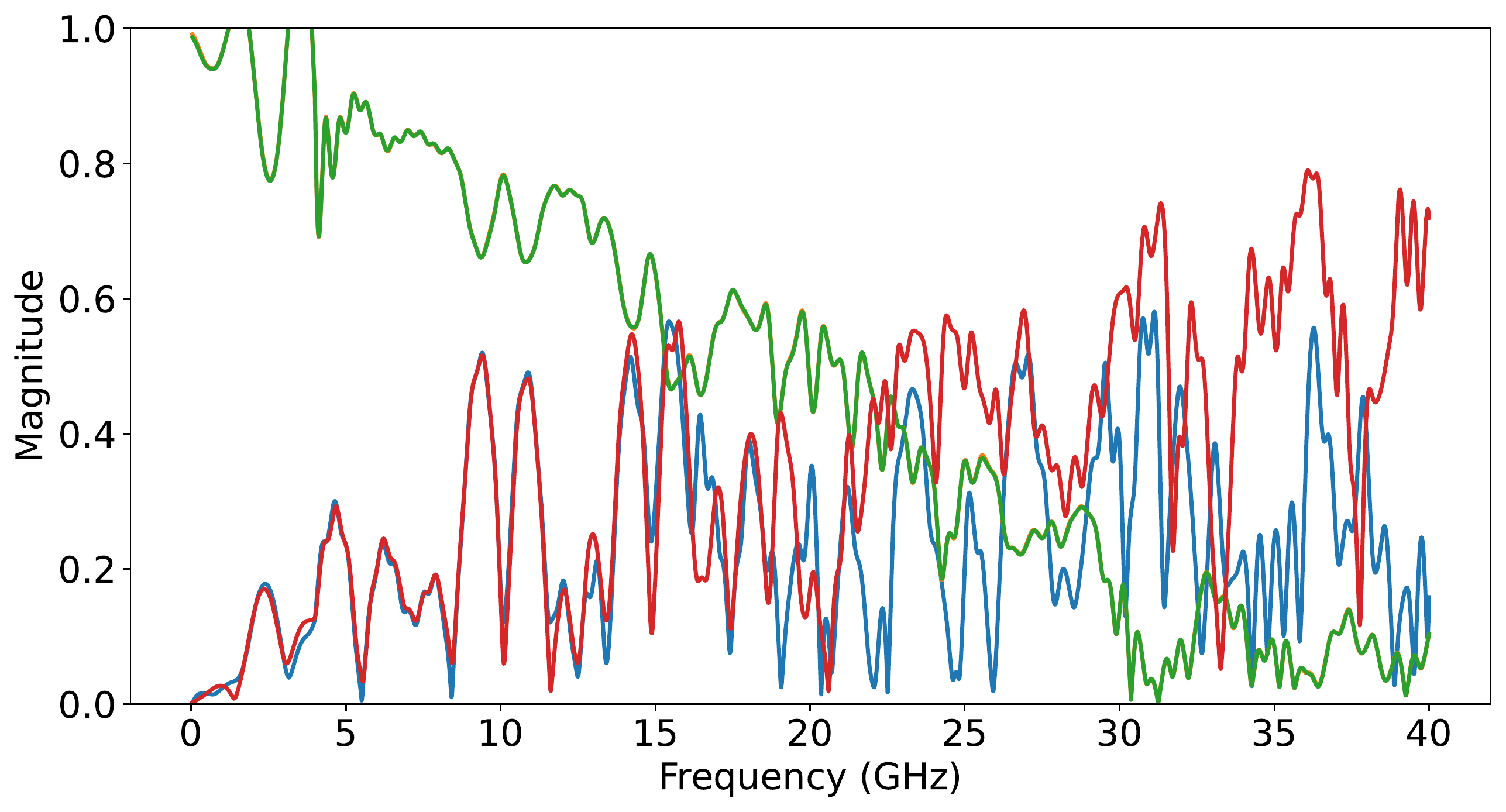} }}%
    \vspace{0.01pt}
    \subfloat[Absolute Error]{{\includegraphics[width=0.49\linewidth]{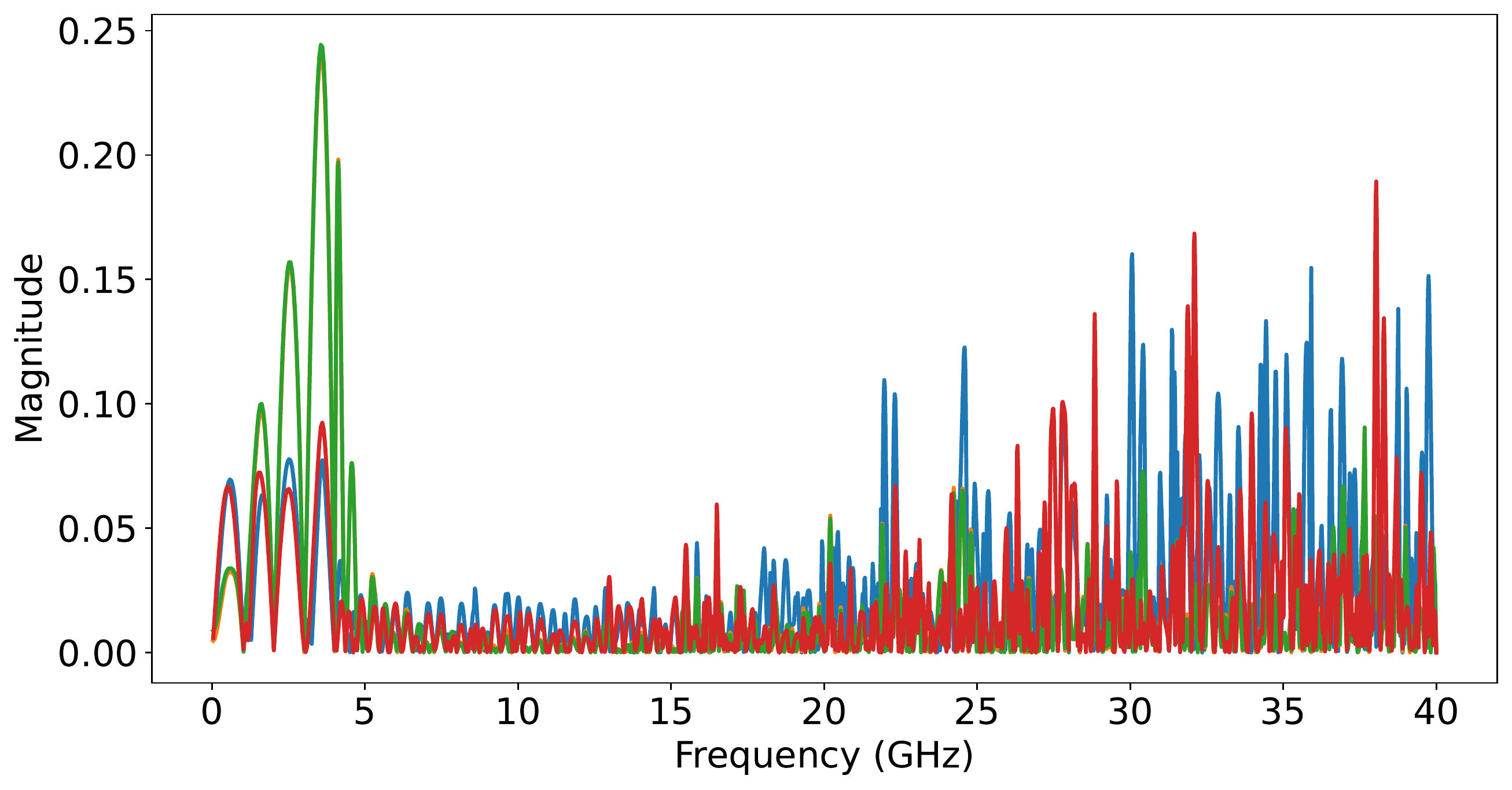} }}%
    \subfloat[Standard Deviation of DIP Output]{{\includegraphics[width=0.49\linewidth]{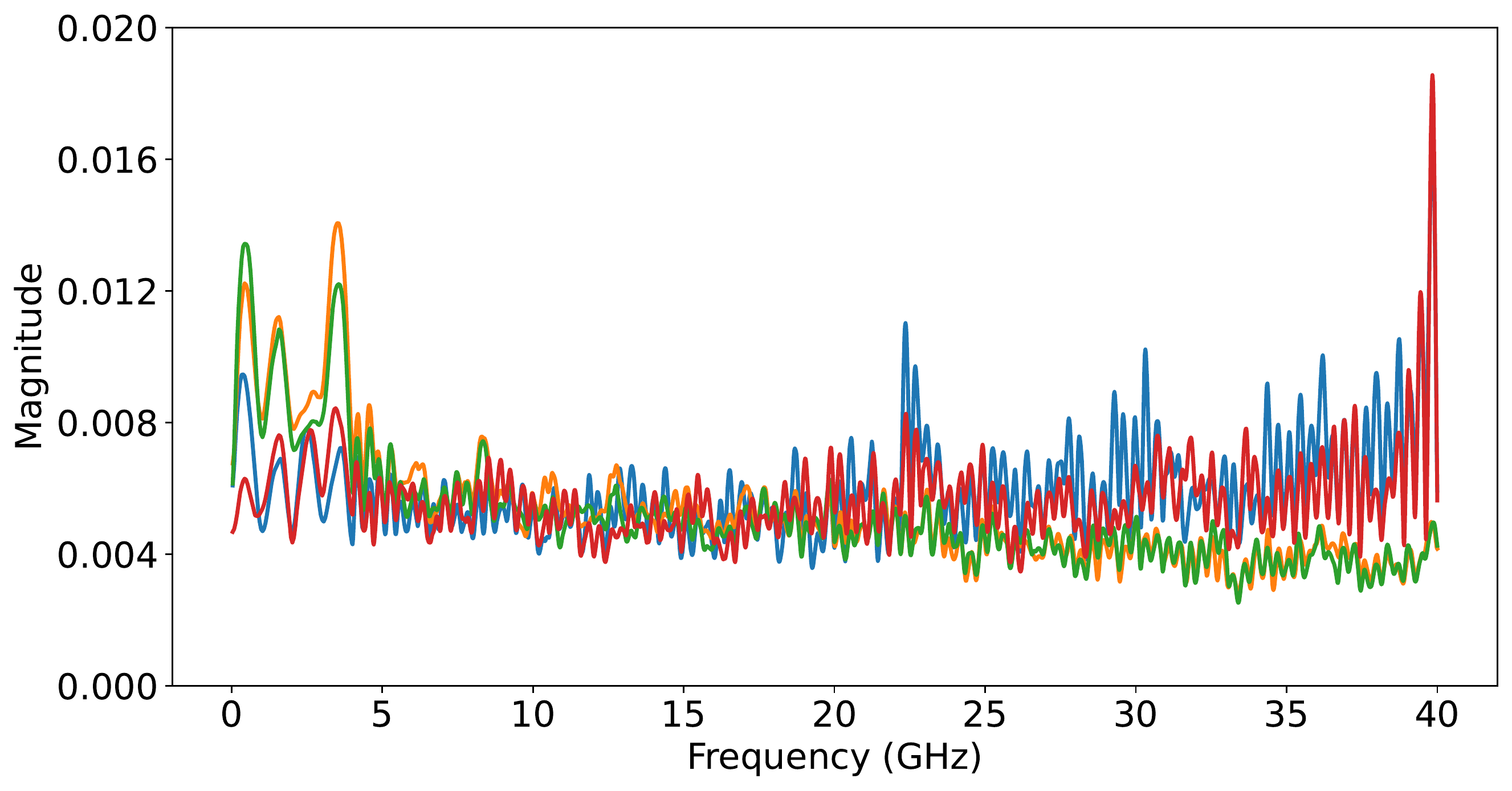} }}%
    \caption{  \textbf{Uncertainty Quantification with SGLD.} We plot the mean and standard deviation of 50 posterior samples from our DIP method using SGLD for fitting a 2-port system. The observed measurements include 104 out of 2081 (about $5\%$) of the available frequency samples. We plot all 4 S-parameters in each graph. Spikes in the standard deviation of DIP outputs predict regions with high error reasonably well. }
    % \vspace{0.1in}
    \label{fig:uncertainty}
\end{figure*}

We propose a novel regularization method that penalizes non-natural frequency domain signals to augment the structural prior provided by DIP. We notice that the frequency response of broadband linear systems are comprised of many nearly-parabolic regions. To investigate a solution, we first turn to classical methods for fitting curves with polynomials. For an unknown scalar-valued function $f(x)$ with known input-output samples $\{x_i, y_i\}_{i=0}^{N-1}$, smoothing splines fit a model function $h(\cdot)$ as
\begin{equation}
\label{eq:smoothing_spline}
    h^* = \argmin_{h} \bigg[ \sum_{i=0}^{N-1} (y_i - h(x_i))^2 + \lambda \int_x \bigg( \frac{d^{2}h(x)}{dx^2} \bigg)^2 dx \bigg],
\end{equation}
where $\lambda$ is a non-negative \emph{smoothing} hyperparameter \cite{hastie2009elements}. The first term is a data-fitting loss similar to the loss in Eq.~\eqref{eq:dip}. The second term is a \emph{smoothness penalty} that penalizes the sum of the squares of the second derivative of the model $h(\cdot)$. This penalty encourages functions to have uniformly small coefficients on their second derivatives, preferring flat quadratic regions interspersed with linear or constant regions. The hyperparameter $\lambda$ controls the intensity of the regularization, with $\lambda = 0$ admitting any function that fits the data and $\lambda = \infty$ allowing only functions with second derivative uniformly equal to zero.  

We propose the following regularization penalty:
\begin{equation}
\label{eq:our_reg}
    R(f(\cdot), \lambda) = \lambda \sum_{i=0}^{p-1} \sum_{j=0}^{p-1} \int_{\omega} \bigg|\frac{d^3f(\omega)[i,j]}{d\omega^3}\bigg| d\omega.
\end{equation}
In other words, we treat each of the $p^2$ S-parameters in a system as separate frequency-domain curves and penalize the absolute value of their third derivatives. By penalizing the third derivatives instead of second, we allow for parabolic regions with arbitrary magnitude. Also, by penalizing the sums of absolute values, we allow for sparse cubic regions linking the parabolic regions instead of cubic regions with uniformly small magnitude as would be the case if we use sums of squares. In practice, we use discrete sums and third-order differences to calculate the integral and third derivative in Eq.~\eqref{eq:our_reg}, respectively.    

\subsection{Training}
\label{sec:training}

Our initial setup given in Eq.~\eqref{eq:our_method} is a minimization problem, which as we detail in Section~\ref{sec:dip} poses the risk of overfitting to the measurements. We therefore perform SGLD as our update algorithm. Our update rule is given by 
\begin{equation}
\begin{aligned}
\label{eq:our_sgld}
        \Delta_{\btheta} &= \frac{\alpha}{2} \bigg[ \nabla_{\btheta} \norm{\Y - \cA(G_{\btheta}(\Z))}_2^2 + \nabla_{\btheta} R(G_{\btheta}(\Z), \lambda) \bigg] + \mathbf{n}, \\ 
    \mathbf{n} &\sim \cN(\mathbf{0}, \alpha \I),
\end{aligned}
\end{equation}
where $\alpha$ is the step size and the first (data-fitting) term represents the likelihood $p(\Y|\btheta)$ while the second (regularization) term is the prior $p(\btheta)$. SGLD allows us to quantify the uncertainty of our method by calculating a sample-wise mean and variance for reconstructions $\hat{\X}_1, \dots, \hat{\X}_n$ drawn i.i.d. from the posterior. 

Following prior works, we do not optimize the input to the network $\Z$ along with the weights. In order to take advantage of the U-Net structure of the network, we initialize the latent as $\Z = \cA^{\dagger}(\Y) \in \R^{r \times f}$ where $\cA^\dagger$ is the adjoint of the forward operator $\cA$. This transformation essentially fills the unknown frequency points in $\Y$ with zeros, ``lifting'' it back into the same dimensions as $\X$. Also following prior work, we add white Gaussian noise to $\Z$, where for iteration $t$ the actual input to the network is given by $\Z_t$ and
\begin{equation}
\label{eq:input_noise}
    \Z_t = \Z + \mathbf{n}_t, \:\: \mathbf{n}_t \sim \cN(\mathbf{0}, \sigma^2_t \I),
\end{equation}
with $\sigma^2_t$ decreases geometrically as $t$ increases.

\begin{figure}[!t]
    \centering
    \includegraphics[width=0.9\linewidth]{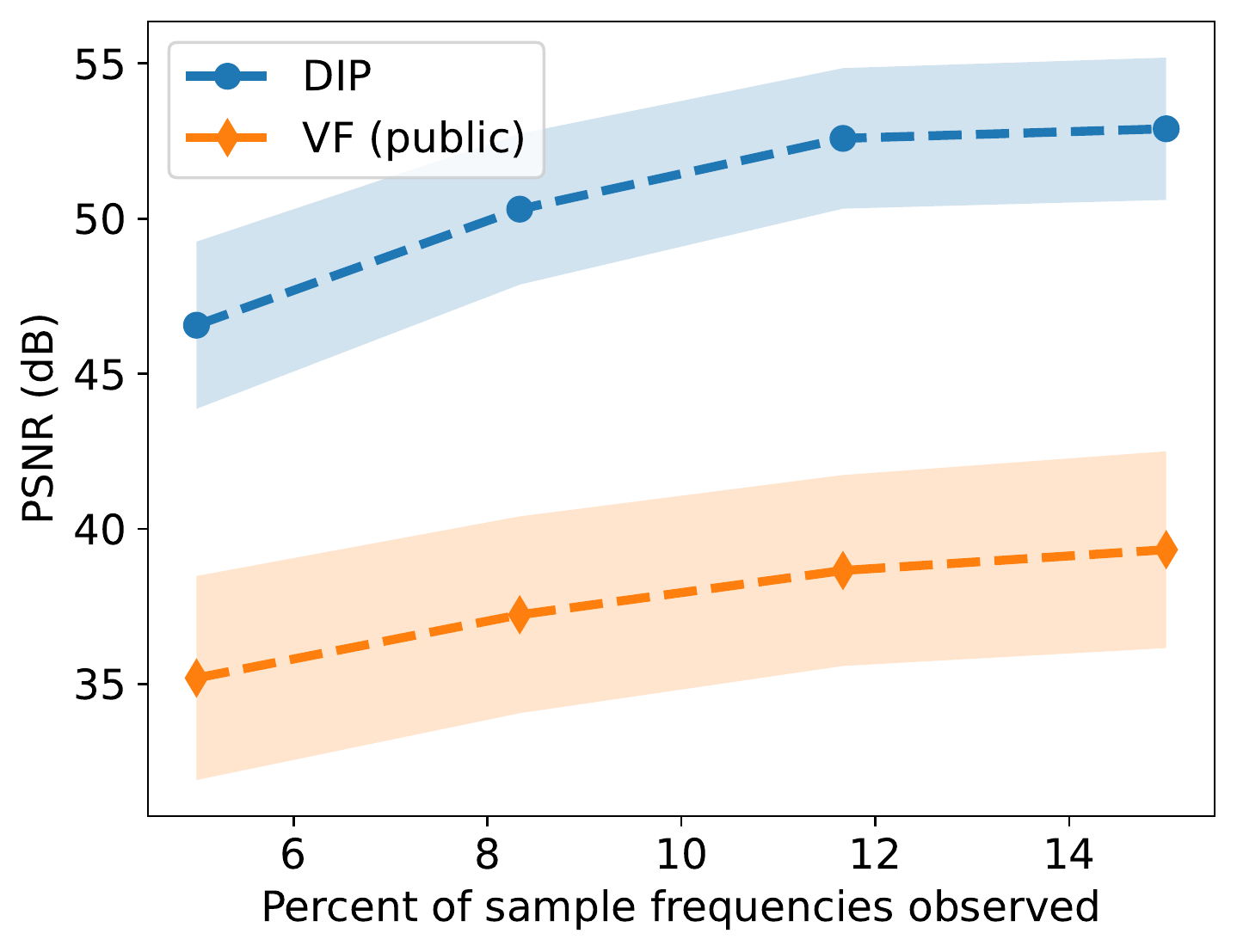}
    \caption{\textbf{Curve Fitting Results for Various Sub-sampling Rates.} We sub-sample simulated S-parameter data at various rates and perform curve fitting on the resulting measurements. We compare DIP to a public implementation of VF and report the mean Peak Signal-to-Noise Ratio (PSNR) between the fitted curves and the fully-sampled data. The shaded areas indicate one standard deviation over different test examples.}
    \label{fig:quant_skrf}
\end{figure}

\section{Experiments}
\label{sec:experiments}

We experiment on simulated S-parameter data with very long electrical channels, which present a more challenging curve fitting problem than shorter channels due to their closely-packed resonances. We create a dataset of 15 simulated frequency curves representing EM solver-generated S-parameters for electrically long interconnects, with up to 16 ports and 9,999 query frequencies. We present the specifications of the examples in our dataset in Table~\ref{tab:specs}.  

As baselines, we compare to the public Scikit-RF implementation of VF \cite{skrf} as well as a proprietary VF implementation from Siemens, used in current industry workflows \cite{hlas_website}. We compare to the public implementation for curve fitting given a fixed, pre-determined number of sub-sampled frequencies. The proprietary VF performs curve fitting using active learning by iteratively fitting available samples, then requesting more frequency samples from the simulator, repeating the process until a stopping criterion is reached. We compare DIP and the public VF to the proprietary VF by selecting an equally-spaced set of measurements with the same sub-sampling ratio as the final set of measurements used by proprietary VF. Although the exact measurements used by DIP and public VF may differ from those used by proprietary VF, keeping the sub-sampling ratio the same means that we do not unfairly advantage or disadvantage any of the methods. 

For all of our experiments, we perform SGLD for 20,000 iterations using a learning rate of $\alpha = 2\times10^{-4}$ and regularization hyperparameter $\lambda = 0.1$. For the additive input noise $\mathbf{n}_t$, we set the starting variance $\sigma_0^2 = 1\times10^{-2}$ and the final variance $\sigma_{20,000}^2 = 1\times10^{-6}$. We set the number of convolution filters in all layers of the network, except for the decoder in the output layer, equal to $\textup{round}(25\sqrt{r})$, where round($\cdot$) denotes a rounding operation. In addition, we set the network depth $N = \textup{ceil}(\log_2 f) - 4$, where ceil($\cdot$) rounds up to the next highest integer. We set the number of filters to $r/2$ in the final decoder in the output layer.      

\subsection{Quantitative Results}
\label{sec:quantitative}

In the first experiment, we create equally-spaced measurements of each of the examples in our dataset using between 5 and 15\% of the available samples. We perform curve-fitting on the measurements using our method and the public implementation of VF, and report the peak signal-to-noise ratio (PSNR) with respect to the fully-sampled reference example. We show a plot of the results in Figure~\ref{fig:quant_skrf}. Our method outperforms public VF in terms of mean PSNR for every sub-sampling rate we test, achieving an improvement of between 12 and 15 dB over VF depending on the rate. The mean PSNR achieved by our method for $5\%$ of samples observed is higher than that of VF for $15\%$ by over $6$ dB, demonstrating that our method can exceed the reconstruction ability of publicly-available curve fitting tools using less than $\frac{1}{3}$ of the samples. Our method also displays less variance than VF in its reconstruction PSNR between examples.

Next, we compare our method and the public VF to proprietary VF using the method we describe in Section~\ref{sec:experiments}. We present our method and public VF with a different number of equally-spaced measurements for each example, depending on the number of measurements used by proprietary VF at the end of its active learning. We present the PSNR results for each method in Table~\ref{tab:results}. Our method achieves the best PSNR in 4 out of 15 cases, and second best in 7 of the 11 cases where proprietary VF performs better. These results demonstrate that our method can perform similarly to, and in some difficult cases, better than handcrafted curve-fitting priors that have been developed and improved for decades.        

\subsection{Qualitative Results}
\label{sec:qualitative}

We display visual curve fitting results from each method for example 0 in Figure~\ref{fig:qualitative_9}. Example 0 presents a particularly challenging curve-fitting problem for VF, having a rapidly-changing magnitude spectrum across all frequency bands. As indicated in Table~\ref{tab:results}, however, our method is able to achieve good results on this example, outperforming the next best method by over 12 dB. The reconstructed curves in Figure~\ref{fig:qualitative_9} tell a similar story. While proprietary VF captures the general shape of the magnitude response, it often produces artifacts such as the extraneous lobes at the high frequencies in the bottom left plot of Figure~\ref{fig:qualitative_9}. Our method, on the other hand, tracks the shape of the S-parameters well without producing noticeable artifacts. Public VF displays poor performance on this example.

As a simpler, 2-port case we plot the visual results for example 2 in Figure~\ref{fig:qualitative_20}. Proprietary VF outperforms our method on this example in terms of PSNR, but only by about 2 dB. This result is impressive, given that proprietary VF is able to choose its samples using active learning and our method is given pre-selected measurements that may not be optimal. As seen in Figure~\ref{fig:qualitative_20}, our method achieves similar qualitative results to proprietary VF.   

\subsection{Uncertainty Quantification}
\label{sec:uncertainty}

One of the benefits of SGLD is that we can draw multiple samples from the posterior and quantify the uncertainty using the sample-wise standard deviation. In this experiment, we draw multiple posterior samples for curve fitting on example 4 with $5\%$ of the frequency samples observed as measurements. Following the authors of \cite{dip_bayesian}, we use a ``burn-in'' period for SGLD, where we only start tracking the sampled SGLD outputs after 15,000 iterations, and track the DIP output every 100 iterations during the post burn-in period for a total of 50 SGLD samples. We plot the mean and standard deviation outputs from our method, along with the reference fully-sampled image and the magnitude of the error in Figure~\ref{fig:uncertainty}. 

The standard deviation plot shows spikes around 0-5 GHz, 20-25 GHz, and 35-40 GHZ, all of which are reflected by corresponding spikes in the error plot at the same frequency ranges. While the magnitude of the standard deviation is around $10\times$ smaller than that of the true error, the modes in the sample-wise standard deviation predict the modes of the absolute error reasonably well. Despite never being trained, our method is able to provide a useful estimate of uncertainty in its predictions.

\subsection{Ablation Study}
\label{sec:ablation}

\begin{figure}[!t]
    \centering
    \includegraphics[width=0.9\linewidth]{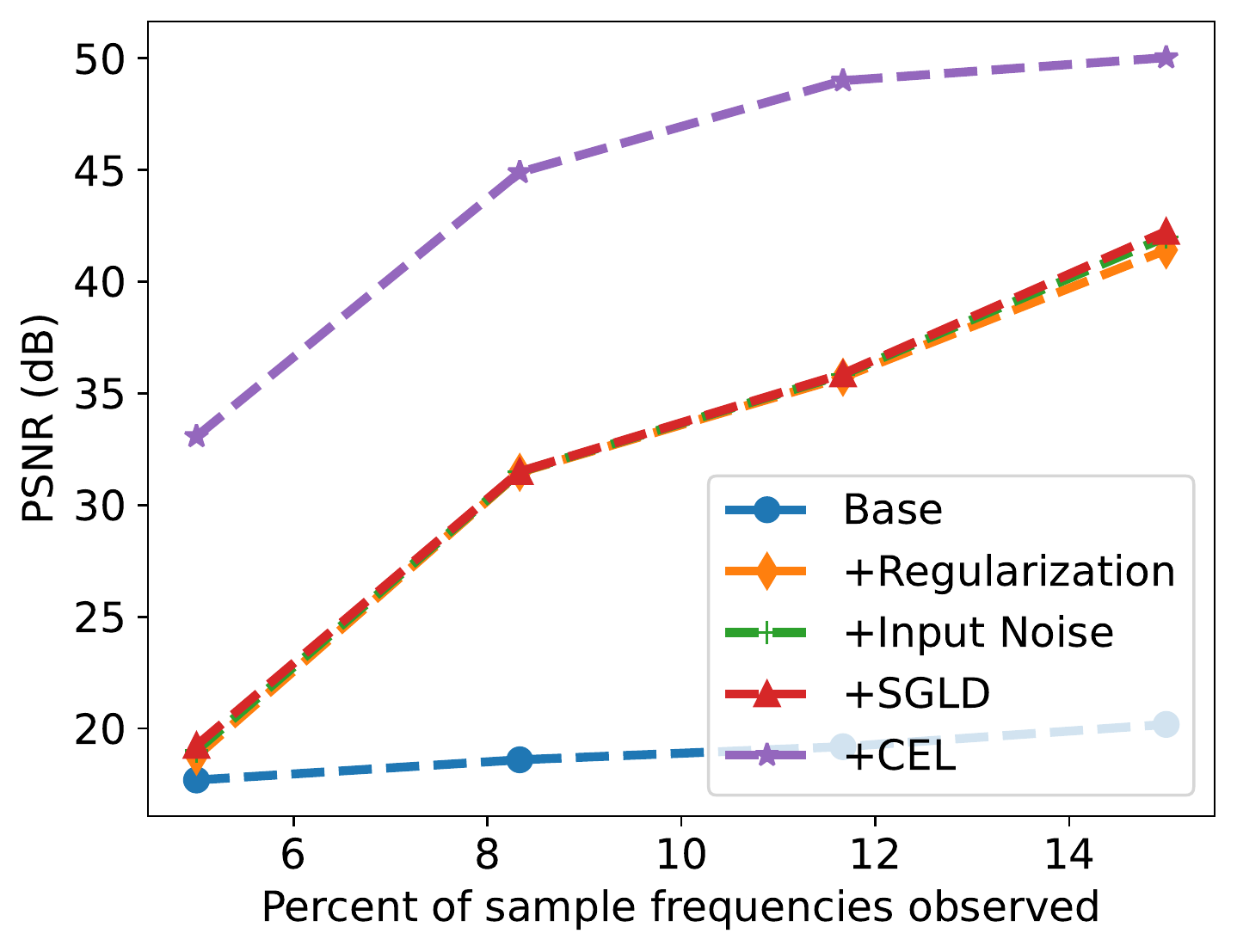}
    \caption{\textbf{Ablation of our Design Choices.} We perform curve fitting on a 4-port system with varying sub-sampling rates using multiple versions of DIP. We start from a ``vanilla'' base DIP model and create the different DIP versions by adding in our proposed regularization, additive noise schedule on the network input, SGLD in place of gradient descent, and a causality enforcement layer (CEL) at the output of the network, in that order. Each additional design choice improves the results, with the regularization and the CEL offering the greatest improvements. }
    \label{fig:ablation}
\end{figure}

Finally, we test the incremental benefits of our design choices by performing an ablation study. We start with a base DIP model that does not use a CEL at the output and simply optimizes the objective in Eq.~\eqref{eq:our_method} using gradient descent with a fixed input $\Z$. One at a time, we re-introduce the regularization term from Eq.~\eqref{eq:our_reg}, the additive input noise schedule $\mathbf{n}_t$, SGLD in place of gradient descent, and the CEL, in that order. We perform curve fitting on example 0 with varying sub-sampling rates using each of these incremental models and plot the results in terms of PSNR in Figure~\ref{fig:ablation}. 

Adding the regularization term significantly improves results, especially for sub-sampling rates higher than 5\%. The additions of input noise and SGLD offer small improvements, while further adding the CEL on top offers another large leap in performance. Overall, our design choices, and particularly our novel regularization, demonstrate a definitive improvement over na\"ive application of DIP.        

\section{Conclusion}
\label{sec:conclusion}

We presented a novel method to extrapolate a detailed frequency response from a limited sampling of the required data using the framework of Deep Image Prior. Our method significantly outperforms a publicly available Vector Fitting implementation, and shows promise versus a proprietary commercial implementation.
Our method uses a novel smoothing regularization term that penalizes non-natural frequency responses and also enforces causality, both incorporated into a DIP optimization framework. 

%nothing good comes out of this
%At present, our method requires no \emph{a priori} system modeling but future models may be able to benefit from additional domain knowledge.

An important future direction for our work is to incorporate active learning into our DIP framework. Active learning methods select which frequency response points to observe in a sequential manner. 
This would obviate the burden of selecting equally-spaced measurements \emph{a priori} to fit with DIP by allowing the active learning algorithm to query the simulator in an online fashion. Since SGLD allows us to draw multiple samples from the posterior distribution, we could use the sample-wise variance at each predicted frequency point to inform our active measurement selection process.   

\bibliography{main}

\begin{thebibliography}{10}

\bibitem{Fast-solvers}
J.~Morsey, B.~Rubin, L.~Jiang, L.~Shan, L.~Eisenberg, D.~Becker, and
  M.~Arseneault, ``The use of fast integral equations solvers for practical
  package and interconnect analysis,'' in {\em 2006 IEEE Electrical Performane
  of Electronic Packaging}, pp.~335--338, IEEE, 2006.

\bibitem{Parallel}
S.~Zuo, D.~G. Do{\~n}oro, Y.~Zhang, Y.~Bai, and X.~Zhao, ``Simulation of
  challenging electromagnetic problems using a massively parallel finite
  element method solver,'' {\em IEEE Access}, vol.~7, pp.~20346--20362, 2019.

\bibitem{AWE-Pade}
T.~West and V.~Jandhyala, ``A pade/spl acute/via awe fast frequency sweep for
  quasi-static coupled electromagnetic and circuit simulation,'' in {\em IEEE
  Antennas and Propagation Society Symposium, 2004.}, vol.~4, pp.~3948--3951,
  IEEE, 2004.

\bibitem{vector_fitting}
B.~Gustavsen and A.~Semlyen, ``Rational approximation of frequency domain
  responses by vector fitting,'' {\em IEEE Transactions on Power Delivery},
  vol.~14, no.~3, pp.~1052--1061, 1999.

\bibitem{dip}
D.~Ulyanov, A.~Vedaldi, and V.~Lempitsky, ``Deep image prior,'' in {\em
  Proceedings of the IEEE conference on computer vision and pattern
  recognition}, pp.~9446--9454, 2018.

\bibitem{vector_fit_pole_relocation}
B.~Gustavsen, ``Improving the pole relocating properties of vector fitting,''
  {\em IEEE Transactions on Power Delivery}, vol.~21, no.~3, pp.~1587--1592,
  2006.

\bibitem{vector_fit_passivity}
B.~Gustavsen and A.~Semlyen, ``Enforcing passivity for admittance matrices
  approximated by rational functions,'' {\em IEEE Transactions on Power
  Systems}, vol.~16, no.~1, pp.~97--104, 2001.

\bibitem{vf_complexity}
A.~Chinea and S.~Grivet-Talocia, ``On the parallelization of vector fitting
  algorithms,'' {\em IEEE Transaction on Components, Packaging and
  Manufacturing Technology}, vol.~1, pp.~1761--1773, Nov 2011.

\bibitem{1d_dip}
S.~Ravula and A.~G. Dimakis, ``One-dimensional deep image prior for time series
  inverse problems,'' in {\em 2022 56th Asilomar Conference on Signals,
  Systems, and Computers}, pp.~1005--1009, 2022.

\bibitem{dip_uncertainty_tomography}
J.~Antorán, R.~Barbano, J.~Leuschner, J.~M. Hernández-Lobato, and B.~Jin,
  ``Uncertainty estimation for computed tomography with a linearised deep image
  prior,'' 2022.

\bibitem{dip_uncertainty_dropout}
M.-H. Laves, M.~T{\"o}lle, and T.~Ortmaier, ``Uncertainty estimation in medical
  image denoising with bayesian deep image prior,'' in {\em Uncertainty for
  Safe Utilization of Machine Learning in Medical Imaging, and Graphs in
  Biomedical Image Analysis: Second International Workshop, UNSURE 2020, and
  Third International Workshop, GRAIL 2020, Held in Conjunction with MICCAI
  2020, Lima, Peru, October 8, 2020, Proceedings 2}, pp.~81--96, Springer,
  2020.

\bibitem{dip_uncertainty_mean_field}
M.~T{\"o}lle, M.-H. Laves, and A.~Schlaefer, ``A mean-field variational
  inference approach to deep image prior for inverse problems in medical
  imaging,'' in {\em Proceedings of the Fourth Conference on Medical Imaging
  with Deep Learning} (M.~Heinrich, Q.~Dou, M.~de~Bruijne, J.~Lellmann,
  A.~Schläfer, and F.~Ernst, eds.), vol.~143 of {\em Proceedings of Machine
  Learning Research}, pp.~745--760, PMLR, 07--09 Jul 2021.

\bibitem{dip_bayesian}
Z.~Cheng, M.~Gadelha, S.~Maji, and D.~Sheldon, ``A bayesian perspective on the
  deep image prior,'' {\em 2019 IEEE/CVF Conference on Computer Vision and
  Pattern Recognition (CVPR)}, pp.~5438--5446, 2019.

\bibitem{ronneberger2015u}
O.~Ronneberger, P.~Fischer, and T.~Brox, ``U-net: Convolutional networks for
  biomedical image segmentation,'' in {\em Medical Image Computing and
  Computer-Assisted Intervention--MICCAI 2015: 18th International Conference,
  Munich, Germany, October 5-9, 2015, Proceedings, Part III 18}, pp.~234--241,
  Springer, 2015.

\bibitem{kkr}
J.~S. Toll, ``Causality and the dispersion relation: Logical foundations,''
  {\em Phys. Rev.}, vol.~104, pp.~1760--1770, Dec 1956.

\bibitem{causal_passive}
H.~M. Torun, A.~C. Durgun, K.~Aygün, and M.~Swaminathan, ``Causal and passive
  parameterization of s-parameters using neural networks,'' {\em IEEE
  Transactions on Microwave Theory and Techniques}, vol.~68, no.~10,
  pp.~4290--4304, 2020.

\bibitem{hastie2009elements}
T.~Hastie, R.~Tibshirani, J.~H. Friedman, and J.~H. Friedman, {\em The elements
  of statistical learning: data mining, inference, and prediction}, vol.~2.
\newblock Springer, 2009.

\bibitem{skrf}
A.~Arsenovic, J.~Hillairet, J.~Anderson, H.~Forstén, V.~Rieß, M.~Eller,
  N.~Sauber, R.~Weikle, W.~Barnhart, and F.~Forstmayr, ``scikit-rf: An open
  source python package for microwave network creation, analysis, and
  calibration [speaker’s corner],'' {\em IEEE Microwave Magazine}, vol.~23,
  no.~1, pp.~98--105, 2022.

\bibitem{hlas_website}
``Hyperlynx advanced solvers.''
\newblock Accessed: 2023-05-22.

\end{thebibliography}
\bibliographystyle{ieeetr}

\end{document}